\documentclass{article}
\usepackage[a4paper]{geometry}
\usepackage{amsmath,amsfonts}
\usepackage{algorithmic}
\usepackage{algorithm}
\usepackage{array}
\usepackage[caption=false,font=normalsize,labelfont=sf,textfont=sf]{subfig}
\usepackage{textcomp}
\usepackage{stfloats}
\usepackage{url}
\usepackage{verbatim}
\usepackage{graphicx}
\usepackage{cite}
\usepackage{hyperref}

\usepackage{numprint}
\usepackage{cleveref}
\usepackage{xspace}
\usepackage{multirow}
\usepackage[x11names]{xcolor}
\usepackage{adjustbox}

\usepackage{todonotes}

\newcommand{\marktext}[2]{\adjustbox{bgcolor=#1}{\mathstrut #2}}

\newcommand{\cotl}{\textsc{Chain of Thought}\xspace}
\newcommand{\stdl}{\textsc{Standard}\xspace}
\newcommand{\std}{\textsc{Std}\xspace}
\newcommand{\cots}{\textsc{CoT}\xspace}
\newcommand{\soall}{\textsc{Spell-out Adjacency List}\xspace}
\newcommand{\soal}{\textsc{SoAL}\xspace}
\newcommand{\gpt}{\marktext{AntiqueWhite1}{\textsc{GPT-4o}}\xspace}
\newcommand{\claude}{\marktext{AntiqueWhite2}{\textsc{Claude-3.7-Sonnet}}\xspace}

\newcommand{\textual}{\marktext{Bisque1}{\textsc{Textual}}\xspace}
\newcommand{\txt}{\marktext{Bisque1}{\textsc{Txt}}\xspace}
\newcommand{\visual}{\marktext{LightCyan1}{\textsc{Visual}}\xspace}
\newcommand{\vis}{\marktext{LightCyan1}{\textsc{Vis}}\xspace}
\newcommand{\slvisual}{\marktext{LightBlue1}{\textsc{Straght-line Visual}}\xspace}
\newcommand{\slv}{\marktext{LightBlue1}{\textsc{SlV}}\xspace}
\newcommand{\slivisual}{\marktext{Turquoise1}{\textsc{Improved Straght-line Visual}}\xspace}
\newcommand{\sliv}{\marktext{Turquoise1}{\textsc{I-SlV}}\xspace}
\newcommand{\orvisual}{\marktext{LightBlue2}{\textsc{Orthogonal Visual}}\xspace}
\newcommand{\orv}{\marktext{LightBlue2}{\textsc{OrV}}\xspace}
\newcommand{\mixed}{\marktext{LavenderBlush2}{\textsc{Mixed}}\xspace}
\newcommand{\mix}{\marktext{LavenderBlush2}{\textsc{Mix}}\xspace}
\newcommand{\slmixed}{\marktext{LightPink1}{\textsc{Straght-line Mixed}}\xspace}
\newcommand{\slm}{\marktext{LightPink1}{\textsc{SlM}}\xspace}

\newcommand{\slimixed}{\marktext{Thistle1}{\textsc{Improved Straght-line Mixed}}\xspace}
\newcommand{\slim}{\marktext{Thistle1}{\textsc{I-SlM}}\xspace}
\newcommand{\ormixed}{\marktext{LightPink2}{\textsc{Orthogonal Mixed}}\xspace}
\newcommand{\orm}{\marktext{LightPink2}{\textsc{OrM}}\xspace}

\begin{document}

%\title{\textsc{GD4LLM}: Evaluating Graph Drawing Techniques for LLMs}
%\title{How Good are LLMs in Understanding Graph Drawings? An Empirical Evaluation}
\title{Graph Drawing for LLMs:\\ An Empirical Evaluation}

\author{Walter~Didimo, 
	Fabrizio Montecchiani, 
	Tommaso~Piselli\\  Department of Engineering, University of Perugia, Italy.\\ \small \{walter.didimo,fabrizio.montecchiani\}@unipg.it,\\  \small tommaso.piselli@dottorandi.unipg.it
    }
    
    \date{}
	
\maketitle

	\begin{abstract}
Our work contributes to the fast-growing literature on the use of Large Language Models (LLMs) to perform graph-related tasks. In particular, we focus on usage scenarios that rely on the visual modality, feeding the model with a drawing of the graph under analysis. We investigate how the model's performance is affected by the chosen layout paradigm, the aesthetics of the drawing, and the prompting technique used for the queries. We formulate three corresponding research questions and present the results of a thorough experimental analysis. Our findings reveal that choosing the right layout paradigm and optimizing the readability of the input drawing from a human perspective can significantly improve the performance of the model on the given task. Moreover, selecting the most effective prompting technique is a challenging yet crucial task for achieving optimal performance.
	\end{abstract}

% make the title area

\section{Introduction}\label{se:introduction}
The landscape of Generative AI expanded tremendously in the last few years, with Large Language Models (LLMs) who have drawn attention due to their strong performance on a wide range of natural language tasks~\cite{LLMSurvey}. Since graphs play a pivotal role in multiple domains, such as recommendation systems and social network analysis~\cite{DBLP:journals/tai/00010YAWP021}, there is an increasing interest in investigating the potential of LLMs on performing graph-related tasks~\cite{DBLP:journals/tkde/JinLHJJH24,DBLP:conf/kdd/RenTYCH24}. Different methods have been proposed to enable LLMs to understand graph structures. One approach consists in  feeding the model with a suitable textual description of the  graph (see, e.g.,~\cite{DBLP:journals/corr/abs-2305-15066}). Alternatively, one can first transform the graph data into a sequence of tokens via specialized modules (such as Graph Neural Networks), and then project this sequence in the LLM's token space (see, e.g.,~\cite{DBLP:journals/corr/abs-2310-05845}). 
In both cases, the assumption is that the graph structure is known as part of the input. 

\begin{figure*}
    \centering
    \includegraphics[width=\linewidth]{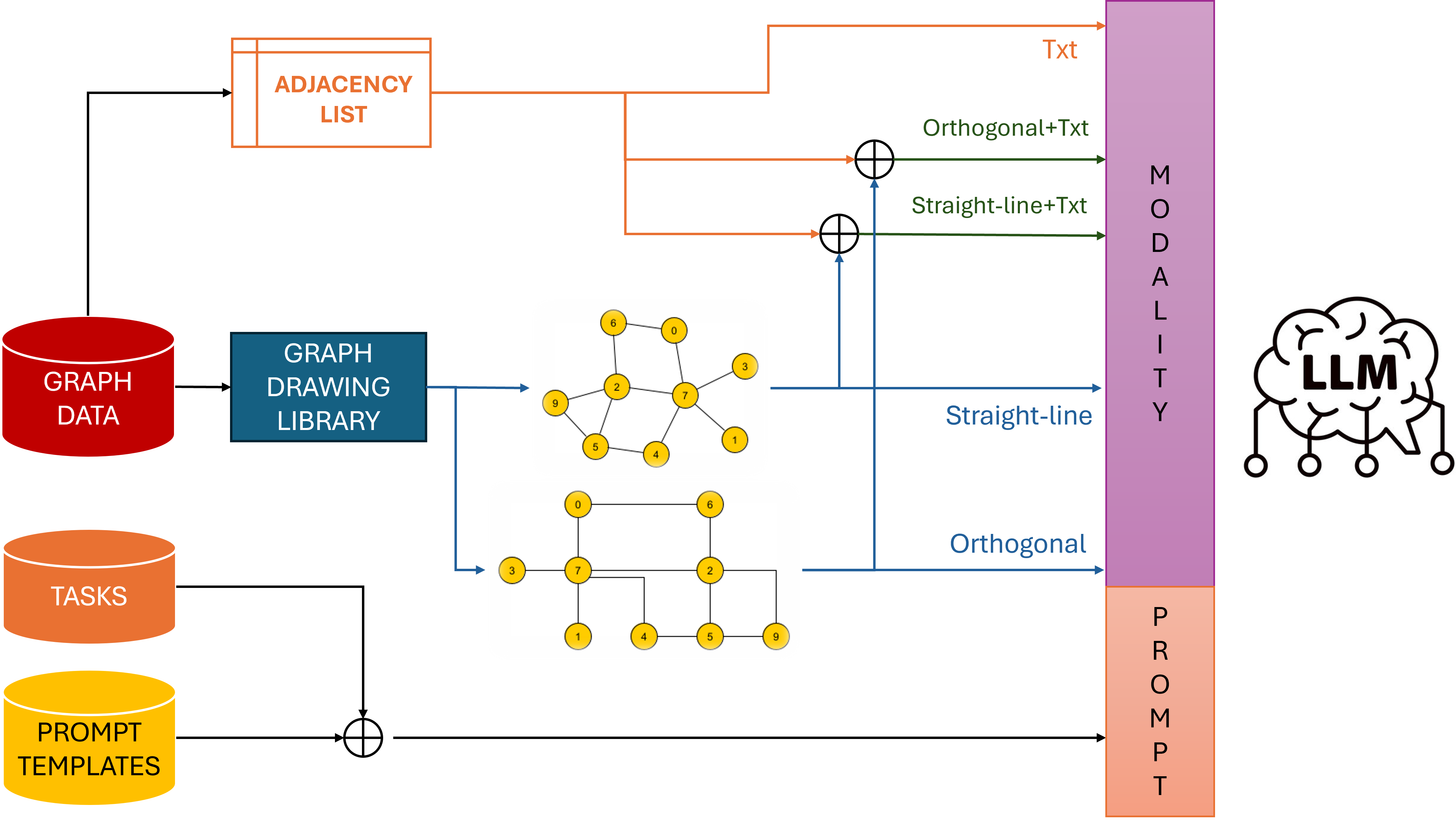}
    \caption{High-level architecture of our experimental framework.}
    \label{fig:enter-label}
\end{figure*}

Despite great efforts on improving graph learning abilities for LLMs, only few studies exploit different modalities other than text. Notably, Das et al.~\cite{DBLP:conf/naacl/DasGSK24} propose an approach based on encoding a graph with multiple modalities, including images and textual motifs, along with suitable prompts. On a similar note, Wei et al.~\cite{DBLP:conf/nips/WeiFJZZWK024} explicitly ask whether incorporating visual information can be beneficial for general graph reasoning, and propose an end-to-end framework integrating visual modality to boost the graph reasoning abilities of LLMs. 

While the two papers above acknowledge the importance of the layout algorithm used to create a visual representation of the graph, they mostly focus on graphic features (e.g., edge thickness), overlooking the impact of different layout paradigms. On the other hand, the graph drawing and network visualization community has always been interested in comparing different layout paradigms in terms of effectiveness on multiple tasks. For instance, Ghoniem et al.~\cite{DBLP:conf/infovis/GhoniemFC04} and Okoe et al.~\cite{DBLP:journals/tvcg/OkoeJK19} compare node-link representations versus matrix-based representations of undirected graphs, whereas Didimo et al.~\cite{DBLP:journals/cgf/DidimoKMT18} compare multiple layout paradigms for directed graphs. See the survey of Burch et al.~\cite{DBLP:journals/access/BurchHWPWH21} for more references. 

Based on the above discussion, our work builds upon the following research questions.  

 \begin{itemize}
    \item {\bf R1} When using the visual modality for graph-related tasks, does the layout paradigm influence the LLM's ability to answer queries on the underlying graph structure?  
    \item {\bf R2} Are there ad-hoc prompting techniques that, paired with a visual representation of the graph, can improve the LLM's performance? 
    \item {\bf R3} Does the quality of the layout, according to human-based readability metrics, impact the LLM's performance? 
\end{itemize}

Besides being of immediate interest for researchers in graph drawing and network visualization, the above questions are also motivated by the following usage scenario. We envision situations in which general-purpose AI assistants support users in solving graph-related tasks, without being integrated with third-party software. As a consequence, the AI assistant can see what the user sees, which is typically a visual representation of the graph computed with a graph layout algorithm, but it may not have access to the underlying graph structure. 

\medskip\noindent{\bf Contribution.} The main contribution of our work can be summarized as follows, and it is motivated by the research questions described above. 

\begin{itemize}
\item Concerning question \textbf{R1}, we empirically evaluate the ability of foundational models in performing graph-related tasks, comparing the textual modality to the visual modality, as well to a mixed modality that exploits both a textual representation and a visual representation of the input graph. Previous experiments tackling similar questions solely focused on \emph{straight-line drawings} as layout paradigm, where nodes are drawn as graphic features (e.g., circles) connected by straight-line segments. Instead, we also consider another popular layout paradigm, namely \emph{orthogonal graph drawing}, where edges are chains of horizontal and vertical segments~\cite{DBLP:books/ph/BattistaETT99,DBLP:reference/crc/DuncanG13,DBLP:conf/dagstuhl/1999dg}. Indeed, orthogonal drawings are widely used for schematic representations in many application domains (e.g., VLSI, software design, database design)~\cite{DBLP:journals/spe/BattistaDPP02,Didimo200635,DBLP:journals/ivs/EiglspergerGKKJLKMS04,DBLP:journals/tc/Valiant81}. 
\item Concerning question \textbf{R2}, we compare multiple prompting techniques. Such techniques are used to craft natural language instructions that provide context or task-specific directions to enhance the efficacy of the model without modifying the core model parameters~\cite{DBLP:journals/corr/abs-2402-07927}. For instance,  \cotl (\cots) is a
technique to trigger a consistent step-by-step reasoning process in LLMs~\cite{DBLP:conf/nips/Wei0SBIXCLZ22}.  Moreover, we introduce and include in the experiments a new technique, which we call \soall (\soal). This technique drives the model through a reasoning strategy in which a preliminary extraction of the adjacency list of the graph from the image is performed, to enhance the downstream task. Our results show that using \soal often leads to good performance, matching prompting techniques  like \cots.
%matching SotA methods such as \cots.
\item Concerning question \textbf{R3}, we run ad-hoc experiments in which we evaluate whether improving the quality of graph layouts according to well-accepted metrics for humans can enhance the ability of LLMs to solve the given tasks. Examples of such metrics are symmetry and number of edge crossings (see, e.g.,~\cite{DBLP:journals/jgaa/PurchaseAC02,DBLP:journals/jea/PurchaseCJ97}). Our experiment supports our hypothesis and paves the way for new research in this direction.
\item Previous experiments on graph-related tasks mostly focus on the fraction of correct answers as a metric to assess the LLM's performance (see, e.g,~\cite{DBLP:journals/corr/abs-2305-15066,DBLP:conf/naacl/DasGSK24,DBLP:journals/corr/abs-2310-05845}). On the other hand, LLMs are prone to \emph{hallucination}, that is, the generation of plausible yet nonfactual content\cite{10.1145/3703155}. In this context, hallucinations may give rise to answers that are syntactically correct (and hence may potentially lead to good accuracy values) but utterly wrong in terms of graph structure. For example, when asked for the length of the shortest path between two vertices of a given graph, the LLM may reply with a correct number, which, however, derives from a path that does not exist in the graph. As an additional contribution, we design specific similarity metrics, which we then use to evaluate the performance of the considered models. Back to the shortest path example, rather than using vanilla accuracy, we  ask the model to spell-out the path and then weigh the accuracy of the answer based on the amount of existing edges in the output path.
\end{itemize}

The remainder of this paper is organized as follows. \Cref{se:related} provides an overview of the research on using LLMs for graph-related tasks. \Cref{se:experiments} forms the core of the paper and is divided into three subsections, each corresponding to one of the experiments conducted to investigate the three research questions outlined above. \Cref{se:conclusions} concludes the paper by summarizing our key findings and discussing the main limitations of our work, as well as the primary research directions it motivates.

\section{Related work}\label{se:related}

The goal of this section is to summarize the main research concerned with the adoption of LLMs for graph-related tasks. We distinguish between black-box models, which do not require any internal change in the LLM, and ad-hoc models, which instead rely on specialized modules that extend the LLM's original architecture. Since our experiments only use foundational models through their publicly available APIs, the former category is the most relevant for our research.

\paragraph{Black-box models} Guo et al.~\cite{DBLP:journals/corr/abs-2305-15066} investigate different text-based graph description languages (e.g., adjacency list and GraphML) combined with several prompting techniques. They consider different structural and semantic tasks on small samples of real-world graphs with few tens of elements. Their analysis suggests that carefully designed graph description languages and prompts have an impact on the achieved performance, which are however still unsatisfactory. Fatemi et al.~\cite{DBLP:conf/iclr/FatemiHP24} perform a similar study on a larger benchmark of small synthetic graphs, focusing only on structural tasks. The experiments reveal that, besides the encoding method, the nature of the graph task and the structure of the input graph all affect the final performance. While the above research mostly deals with the design of innovative encoding schemes and prompting techniques, the size of the considered graphs is limited to the length of the context window. In particular, this constraint represents an intrinsic limit for text-only modalities, which is therefore not suitable for large graphs. Das et al.~\cite{DBLP:conf/naacl/DasGSK24} explore multiple modalities to encode a graph, namely text, images, and motifs. The motif encoding is less verbose than a full textual description of the graph, capturing essential patterns around single nodes and balancing the trade-off between local and global perspective. On the other hand, images require a fixed amount of tokens to convey the whole graph structure and rely on the vision capabilities of recent LLMs. Two key findings extracted from~\cite{DBLP:conf/naacl/DasGSK24}: the image modality gives the best trade-off between number of tokens used to encode the input graph and performance on graph classification tasks; the effectiveness of the image modality on graph classification tasks positively correlates with the human readability of the visualization. This last finding sheds light on the potential impact of readability metrics and layout paradigms on the LLM's ability to understand the underlying graph structure.

\paragraph{Ad-hoc models}  GraphLLM~\cite{DBLP:journals/corr/abs-2310-05845} combines LLMs with graph transformers for graph reasoning tasks. GraphLM+~\cite{DBLP:journals/corr/abs-2403-04483} is a model fine-tuned on a benchmark called GraphInstruct. The benchmark contains small synthetic graphs with textual descriptions and several structural tasks. GraphGPT~\cite{DBLP:conf/sigir/Tang00SSCY024} introduces a text-graph grounding paradigm to align encodings of graph structures with the natural language space and self-supervised instruction tuning; the model is evaluated with medium-size real-world networks on graph learning tasks. Of greater interest for our research is GITA~\cite{DBLP:conf/nips/WeiFJZZWK024}, an end-to-end framework aimed at boosting the graph reasoning abilities of an LLM through the integration of the visual modality. A key finding here is that integrating visual and textual information can indeed lead to increased performance.

\section{Experimental analysis}\label{se:experiments}

In this section, we present the experiments we did in order to investigate the three research questions \textbf{R1}, \textbf{R2}, and \textbf{R3}. \textbf{All experimental data (including benchmarks, drawings, code, and full prompts) are publicly available\footnote{To be provided after publication or under request.}}  For all experiments we exploited the public APIs of the following two multi-modal LLMs:

\begin{itemize}
    \item \gpt: one of the most recent and cost-effective\footnote{See the following leader-board about performance-cost trade-off: \url{https://arcprize.org/leaderboard} } versions of the popular OpenAI's technology~\cite{DBLP:journals/corr/abs-2410-21276}.
    \item \claude: the latest Anthropic's model\footnote{\url{https://www.anthropic.com/news/claude-3-7-sonnet}}, currently showing state-of-the-art performance\footnote{Leading model in April 2025, \url{https://web.lmarena.ai/leaderboard}; see also~\cite{white2025livebench}.}
\end{itemize}

\noindent Since our goal is to evaluate the ability of LLMs to understand graphs rather than generating code, in our prompts we \emph{do not} ask the model to generate any code in order to solve the given task.
We next describe the experiments in detail, grouped by research question.

\subsection{Experiment 1: Comparing multiple drawing paradigms}
This experiment aims to investigate \textbf{R1} under multiple perspectives. 
We begin by describing the experimental set-up, and we continue with a discussion of the results.

\smallskip
\subsubsection{Experimental set-up}
We describe the input modalities, tasks, prompting techniques, and datasets used in our experiment.

\smallskip\noindent{\bf Input modalities.}
We considered three main input modalities.

\begin{figure}
    \centering
    \includegraphics[width=0.8\linewidth]{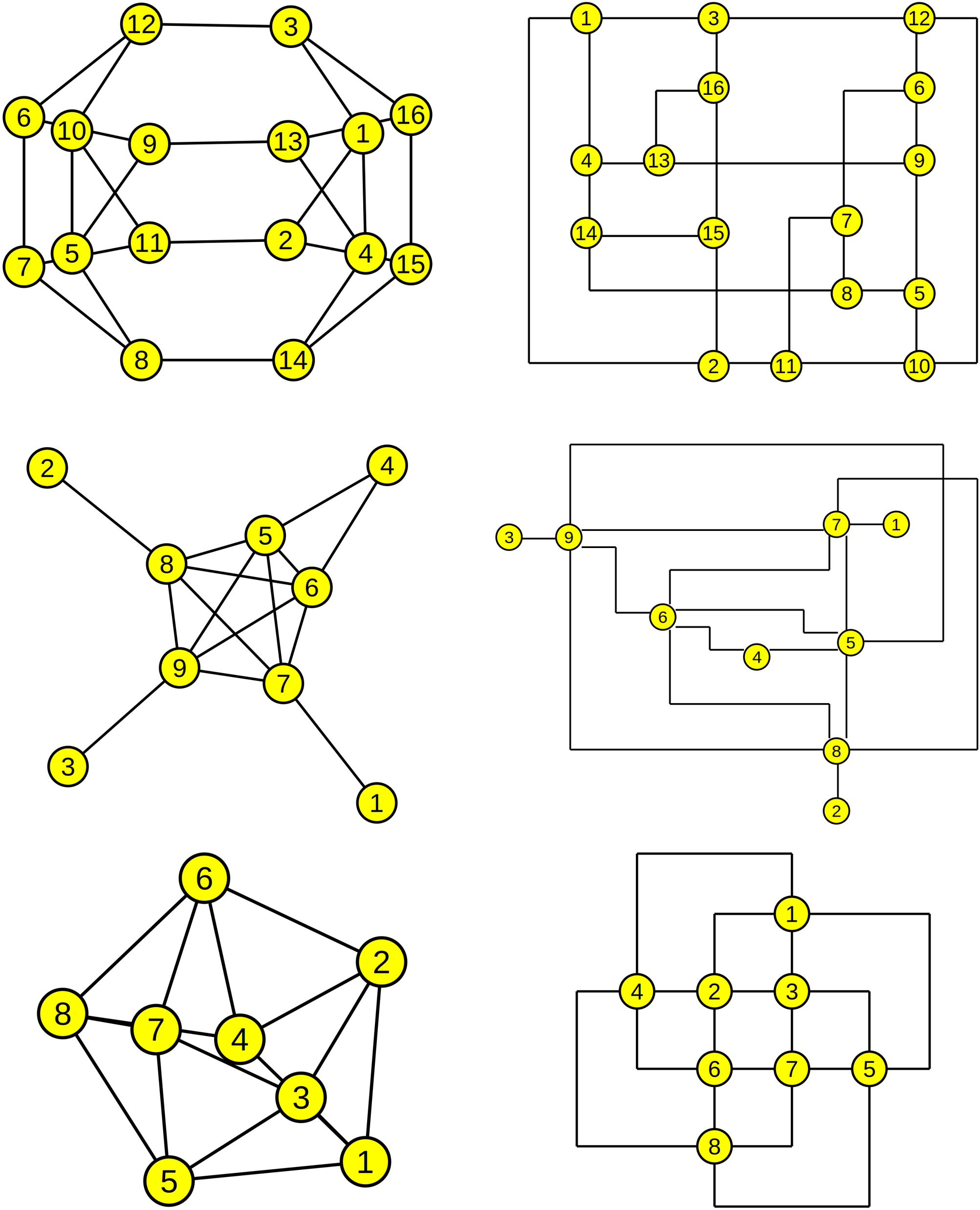}
    \caption{Examples of drawings computed for the \textsc{SlV} (left) and \textsc{OrV} (right) modalities. The first row shows a graph from \textsc{Bench-1}, the second row shows a graph from \textsc{Bench-2} (with a max clique of size five), the third row shows a graph from \textsc{Bench-3} (with a min independent set of size six).}
    \label{fig:drawings}
\end{figure}

\begin{itemize}
    \item  \textual (\txt): a textual description of the input graph in the form of adjacency list. Testing this modality is useful for comparative purposes.
    \item \visual (\vis): an image depicting a drawing of the input graph, without any further information in terms of graph structure. The image resolution is fixed such that the width is $1024$ pixels and the height is scaled based on the drawing's aspect-ratio. This modality comes in two different types, one for each of the two considered graph drawing paradigms.
    \begin{itemize}
        \item \slvisual (\slv): A straight-line drawing of the input graph computed with a force-directed algorithm called FMMM~\cite{DBLP:conf/gd/HachulJ04,DBLP:journals/jgaa/HachulJ07}, available in the OGDF library~\cite{DBLP:reference/crc/ChimaniGJKKM13}. Straight-line drawings are widely adopted due to their intuitiveness. Also, force-directed algorithms are a popular choice for computing straight-line drawings due to their availability, scalability, and flexibility. See \Cref{fig:drawings} for examples of instances used in our experiments.
        \item \orvisual (\orv): An orthogonal drawing of the input graph computed with the implementation available in the OGDF library~\cite{DBLP:reference/crc/ChimaniGJKKM13}. Orhtogonal drawings are commonly used for schematics in light of the high angular and crossing resolution they offer (all angles at nodes and edge crossings are multiples of $90^\circ$). See \Cref{fig:drawings} for examples of instances used in our experiments.
    \end{itemize}
    \item \mixed (\mix): Both the textual and the visual modalities together. We distinguish between \slmixed (\slm) when the visual type is \slv, and \ormixed (\orm) when the visual type is \orv.
\end{itemize}

\smallskip\noindent{\bf Tasks.}
We considered four tasks, which cover different levels of analysis on the graph layout, requiring local and global inspections, and different levels of complexity. These tasks have also been considered in previous experiments (see, e.g.,~\cite{DBLP:journals/corr/abs-2407-00379}).
%We focus on structural tasks of different complexity.
For each task we define an ad-hoc accuracy metric to evaluate the corresponding performance.

\begin{itemize}
%   \item \textsc{Adjacency List} (\textsc{AL}):
   \item \textsc{Common Neighbor} (\textsc{CoNe}): Given two (randomly) selected nodes, we ask how many neighbors they share. This is a relatively simple and local task. To counteract hallucinations, we do not only ask for the numerical value, but also for the list of nodes forming the shared neighborhood. The accuracy is  computed as the Jaccard index between the answered set $A$ and the correct set $B$: $$\alpha_{\text{CoNe}} = \frac{|A \cap B|}{|A \cup B|}.$$

   \item \textsc{Shortest Path} (\textsc{ShPa}): Given two (randomly) selected nodes, we ask for the length of the shortest path between them. This is a more difficult and more global task. To counteract hallucinations, we do not only ask for the numerical value, but also for a candidate path that matches the shortest length. Since there might be exponentially many paths with the same length, we compute the accuracy as follows. Let $\delta$ and $\Delta$ be the outputted and correct length, respectively. Also, let $\sigma$ be the fraction of existing edges in the path outputted by the model. We use the measure: $$\alpha_{\text{ShPa}} = \min{\{\frac{\delta}{\Delta}, \frac{\Delta}{\delta}\}} \times \min{\{\frac{\sigma}{\Delta},\frac{\Delta}{\sigma}\}}.$$ Note that, if the model outputs a correct path, then $\alpha_{\text{ShPa}}=1$. On the other hand, if the model outputs a path with the correct length but in which only half of the edges exist in the graph, then $\alpha_{\text{ShPa}}=0.5$.

   \item \textsc{Max Clique} (\textsc{MaxC}): We ask for the size of the maximum clique in the graph. This is a very difficult (NP-hard) and global task. To counteract hallucinations, we do not only ask for the numerical value, but also for a candidate clique of maximum size. Since there might be exponentially many cliques of the same size, we compute the accuracy as follows. Let $\delta$ and $\Delta$ be the outputted and correct size, respectively. Also, let $\sigma$ be the fraction of existing edges in the clique outputted by the model. We use the measure: $$\alpha_{\text{MaxC}} = \min{\{\frac{\delta}{\Delta}, \frac{\Delta}{\delta}\}} \times \frac{2\sigma}{\Delta(\Delta-1)}.$$ Again, if the model outputs a correct clique, then $\alpha_{\text{MaxC}}=1$, whereas a clique of right size but in which half of the edges do not actually exist leads to $\alpha_{\text{MaxC}}=0.5$ (recall that $\frac{\Delta(\Delta-1)}{2}$ is the number of edges in a clique with $\Delta$ nodes).
   
   \item \textsc{Min Vertex-Cover} (\textsc{MinVC}): We ask for the size of the minimum vertex cover in the graph. This is again a very difficult (NP-hard) and global task. To counteract hallucinations, we do not only ask for the numerical value, but also for a candidate vertex cover of minimum size. Again, there might be exponentially many sets forming a vertex cover of fixed size, hence we compute the accuracy as follows. Let $\delta$ and $\Delta$ be the outputted and correct size, respectively. Also, let $\sigma$ be the fraction of \emph{uncovered} edges for the vertex cover outputted by the model, and let $m$ be the total number of edges of the graph. We use the measure: $$\alpha_{\text{MinVC}} = \min{\{\frac{\delta}{\Delta}, \frac{\Delta}{\delta}\}} \times (1-\frac{\sigma}{m}).$$ Again, if the model outputs a vertex cover of minimum size, then $\alpha_{\text{MinVC}}=1$, while a set of vertices of the right size but that covers only half of the edges leads to $\alpha_{\text{MinVC}}=0.5$.
\end{itemize}

\smallskip\noindent{\bf Prompting techniques.} We started by defining two main prompting techniques.

\begin{itemize}
   \item \stdl (\std): A standard prompt in which the input modality and the task are clearly explained, without any further hint in terms of reasoning strategy. We also adopt common best practices~\cite{DBLP:journals/corr/abs-2402-07927} such as role playing (e.g., ``You are a data scientist...''), length control (e.g., ``Answer with a number in the range [...]'') and chain-of-verification (e.g., ``Before submitting your final answer, verify that...'').
   \item \cotl (\cots): The above standard prompt, paired with a step-by-step reasoning suggestion~\cite{DBLP:conf/nips/Wei0SBIXCLZ22}.
\end{itemize}

\noindent Then, each of the aforementioned prompting technique is combined with the following in-context learning strategies~\cite{DBLP:conf/emnlp/Dong0DZMLXX0C0S24}, that is, strategies to let the model learn from a few examples given as part of
the context.

\begin{itemize}
   \item \textsc{Zero Shots} (\textsc{Zero}): No examples of the given task are given, hence no in-context learning is possible for the model.
   \item \textsc{Few Shots} (\textsc{Few}): two examples are given and encoded with the same input modality, along with a correct answer. If this strategy is paired with the \cots technique, the examples also include a possible step-by-step reasoning strategy tailored to the specific task.
\end{itemize}

\noindent Thus, overall, we have four prompting techniques: \textsc{Std-Zero}, \textsc{Std-Few}, \textsc{CoT-Zero}, \textsc{CoT-Few}.

\smallskip\noindent{\bf Datasets.} We consider the following three benchmarks, which have been designed based on the tasks illustrated before. The benchmarks have been generated using the House of Graphs application~\cite{DBLP:journals/dam/CoolsaetDG23}, which allows to search for graphs satisfying multiple structural properties (e.g., with controlled vertex cover or maximum clique size)\footnote{House of Graphs is a very popular tool, see the following URL for a list of papers using it: \url{https://houseofgraphs.org/publications}}.

\begin{itemize}
   \item \textsc{Graph Benchmark 1} (\textsc{Bench-1}): $20$ graphs, with number of vertices between $6$ and $50$, with different topologies from small planar graphs to more complex graphs with dense communities. This benchmark has been used for tasks \textsc{CoNe} and \textsc{ShPa}.
   
   \item \textsc{Graph Benchmark 2} (\textsc{Bench-2}): $20$ graphs with controlled structure such that the maximum clique size varies in the range $[2,7]$. This benchmark hae been used for task \textsc{MaxC}.
   
   \item \textsc{Graph Benchmark 3} (\textsc{Bench-3}): $20$ graphs with controlled structure such that the minimum vertex cover size varies in the range $[1,26]$. This benchmark has been used for task \textsc{MinVC}.
\end{itemize}

\smallskip\noindent{\bf Further considerations.} For the sake of robustness, for tasks \textsc{CoNe} and \textsc{ShPa}, we always pick two pairs of nodes and average the obtained accuracy. This is of course not possible for tasks \textsc{MaxC} and \textsc{MinVC}, which only take the drawing as input. Moreover, beside accuracy, we measure the latency and cost of the answer in terms of total number of tokens (i.e., input tokens plus output tokens). It is worth remarking that different LLMs adopt different tokenization methods (and pricing policies), therefore the numbers of \gpt and \claude cannot be directly compared. 

\subsubsection{Results}

\newcolumntype{?}{!{\vrule width 1.5pt}}

\begin{table}[h]
\renewcommand{\arraystretch}{1.8}
\scriptsize
    \centering
    \resizebox{0.8\columnwidth}{!}{\begin{tabular}{|c c?c|c|c|c?c|c|c|c|}
    \hline 
    \multicolumn{10}{|c|}{\gpt}\\
  & & \multicolumn{4}{c?}{\textsc{Accuracy $\alpha_{\text{CoNe}}$}}   & \multicolumn{4}{c|}{\textsc{Total Tokens}} \\
  & & \multicolumn{2}{c|}{\textsc{\std}} & \multicolumn{2}{c?}{\textsc{CoT}}  & \multicolumn{2}{c|}{\textsc{\std}} & \multicolumn{2}{c|}{\textsc{CoT}} \\
  & \textbf{Modality}  & \textsc{Zero} &  \textsc{Few} & \textsc{Zero} &  \textsc{Few}  & \textsc{Zero} &  \textsc{Few} & \textsc{Zero} &  \textsc{Few} \\\hline
    & \txt &0.75 &0.83 &\textbf{1.00} &0.98 & \textbf{321} &780 &613  &1\,199  \\\hline
    \parbox[t]{1mm}{\multirow{2}{*}{\rotatebox[origin=c]{90}{\visual}}} & \slv &0.60 &0.63 &0.58 &0.55 &969 &2\,557 &1\,128 &2\,883\\
    & \orv &\textcolor{Coral3}{0.54} &0.60 &0.56 &0.55 &1\,002 &2\,590 &1\,162  &2\,911\\\hline
   \parbox[t]{1mm}{\multirow{2}{*}{\rotatebox[origin=c]{90}{\mixed}}} & \slm &0.83  &0.89 &\textbf{1.00} &\textbf{1.00} &1\,168 &3\,158 &1\,342 &3\,541\\
    & \orm &0.93 &0.86 &\textbf{1.00} &\textbf{1.00} &1\,201 &3\,191 &1\,378 &\textcolor{Coral3}{3\,579}\\
\hline
\multicolumn{10}{|c|}{\claude}\\
  & & \multicolumn{4}{c?}{\textsc{Accuracy $\alpha_{\text{CoNe}}$}}   & \multicolumn{4}{c|}{\textsc{Total Tokens}} \\
  & & \multicolumn{2}{c|}{\textsc{\std}} & \multicolumn{2}{c?}{\textsc{CoT}}  & \multicolumn{2}{c|}{\textsc{\std}} & \multicolumn{2}{c|}{\textsc{CoT}} \\
  & \textbf{Modality}  & \textsc{Zero} &  \textsc{Few} & \textsc{Zero} &  \textsc{Few}  & \textsc{Zero} &  \textsc{Few} & \textsc{Zero} &  \textsc{Few} \\\hline
    & \txt &0.95 &0.94 &\textbf{1.00} &\textbf{1.00} &\textbf{331} &791 &\,647  &1\,235  \\\hline
    \parbox[t]{1mm}{\multirow{2}{*}{\rotatebox[origin=c]{90}{\visual}}}& \slv &0.62 &0.35 &0.62 &0.62 &1\,540 &4\,298 &1\,882 &4\,701\\
    & \orv &0.73 &\textcolor{Coral3}{0.35} &0.78 &0.82 &1\,509 &4\,264 &1\,852  &4\,686\\\hline
   \parbox[t]{1mm}{\multirow{2}{*}{\rotatebox[origin=c]{90}{\mixed}}} & \slm &0.97  &0.95 &\textbf{1.00} &0.98 &1\,739 &4\,900 &2\,079 &5\,403\\
    & \orm &0.98 &0.96 &\textbf{1.00} &0.99 &1\,711 &4\,869 &2\,060 &\textcolor{Coral3}{5\,365}\\\hline
    \end{tabular}}
    \vspace{1mm}
    \caption{Experiment 1: Performance on task \textsc{Common Neighbor}. \\Best (worst) values in \textbf{bold} (\textcolor{Coral3}{red}).}
    \label{tab:cn}
\end{table}

\newcolumntype{?}{!{\vrule width 1.5pt}}
\begin{table}[h]
\renewcommand{\arraystretch}{1.8}
\scriptsize
    \centering
    \resizebox{0.8\columnwidth}{!}{\begin{tabular}{|c c?c|c|c|c?c|c|c|c|}
    \hline 
    \multicolumn{10}{|c|}{\gpt}\\
  & & \multicolumn{4}{c?}{\textsc{Accuracy}  $\alpha_{\text{ShPa}}$}   & \multicolumn{4}{c|}{\textsc{Total Tokens}} \\
  & & \multicolumn{2}{c|}{\textsc{\std}} & \multicolumn{2}{c?}{\textsc{CoT}}  & \multicolumn{2}{c|}{\textsc{\std}} & \multicolumn{2}{c|}{\textsc{CoT}} \\
  & \textbf{Modality}  & \textsc{Zero} &  \textsc{Few} & \textsc{Zero} &  \textsc{Few}  & \textsc{Zero} &  \textsc{Few} & \textsc{Zero} &  \textsc{Few} \\\hline
    & \txt &0.86 &0.96 &0.98 &0.95 &\textbf{332} &794 &1\,340  &2\,034  \\\hline
    \parbox[t]{1mm}{\multirow{2}{*}{\rotatebox[origin=c]{90}{\visual}}} &\slv &0.49 &\textcolor{Coral3}{0.47} &0.55 &0.52 &982 &2\,573 &1\,365 &3\,186\\
    & \orv &0.71 &0.65 &0.69 &0.70 &1\,020 &2\,609 &1\,376  &3\,217\\\hline
   \parbox[t]{1mm}{\multirow{2}{*}{\rotatebox[origin=c]{90}{\mixed}}} & \slm &0.83  &0.90 &\textbf{1.00} &0.93 &1\,181 &3\,174 &1\,535 &3\,843\\
    & \orm &0.93 &0.91 &0.97 &0.97 &1\,216 &3\,209 &1\,564 &\textcolor{Coral3}{3\,866}\\
    \hline 
    \multicolumn{10}{|c|}{\claude}\\
     & & \multicolumn{4}{c?}{\textsc{Accuracy} $\alpha_{\text{ShPa}}$}   & \multicolumn{4}{c|}{\textsc{Total Tokens}} \\
  & & \multicolumn{2}{c|}{\textsc{\std}} & \multicolumn{2}{c?}{\textsc{CoT}}  & \multicolumn{2}{c|}{\textsc{\std}} & \multicolumn{2}{c|}{\textsc{CoT}} \\
  & \textbf{Modality}  & \textsc{Zero} &  \textsc{Few} & \textsc{Zero} &  \textsc{Few}  & \textsc{Zero} &  \textsc{Few} & \textsc{Zero} &  \textsc{Few} \\\hline
    & \txt &0.93 &0.93 &0.98 &\textbf{1.00} &\textbf{342} &805 &944  &1\,537  \\\hline
    \parbox[t]{1mm}{\multirow{2}{*}{\rotatebox[origin=c]{90}{\visual}}} &\slv &\textcolor{Coral3}{0.64} &0.65 &0.68 &0.68 &1\,551 &4\,310 &2\,025 &5\,014\\
    & \orv &0.82 &0.74 &0.88 &0.89 &1\,525 &4\,284 &2\,004  &4\,992\\\hline
   \parbox[t]{1mm}{\multirow{2}{*}{\rotatebox[origin=c]{90}{\mixed}}} & \slm &0.97  &0.88 &0.99 &0.96 &1\,751 &4\,912 &2\,354 &\textcolor{Coral3}{5\,788}\\
    & \orm &0.98 &0.96 &\textbf{1.00} &\textbf{1.00} &1\,723 &4\,884 &2\,324 &5\,748 \\\hline
    \end{tabular}}
    \vspace{1mm}
    \caption{Experiment 1: Performance on task \textsc{Shortest Path}. \\Best (worst) values in \textbf{bold} (\textcolor{Coral3}{red}).}
    \label{tab:sp}
\end{table}

\renewcommand{\arraystretch}{1.8}
\newcolumntype{?}{!{\vrule width 1.5pt}}

\begin{table}[h]
\scriptsize
    \centering
    \resizebox{0.8\columnwidth}{!}{\begin{tabular}{|c c?c|c|c|c?c|c|c|c|}
     \hline 
    \multicolumn{10}{|c|}{\gpt}\\
  & & \multicolumn{4}{c?}{\textsc{Accuracy} $\alpha_{\text{MaxC}}$}   & \multicolumn{4}{c|}{\textsc{Total Tokens}} \\
  & & \multicolumn{2}{c|}{\textsc{\std}} & \multicolumn{2}{c?}{\textsc{CoT}}  & \multicolumn{2}{c|}{\textsc{\std}} & \multicolumn{2}{c|}{\textsc{CoT}} \\
  & \textbf{Modality}  & \textsc{Zero} &  \textsc{Few} & \textsc{Zero} &  \textsc{Few}  & \textsc{Zero} &  \textsc{Few} & \textsc{Zero} &  \textsc{Few} \\\hline
    & \txt &0.80 &0.79 &0.81 &\textbf{0.87} &\textbf{294} &544 & 1\,085  & 1\,621  \\\hline
    \parbox[t]{1mm}{\multirow{2}{*}{\rotatebox[origin=c]{90}{\visual}}} &\slv &0.78 &0.79 &0.85 &0.86 &1\,002 &2\,574 &1\,404 &3\,255\\
    & \orv &0.71 &0.71 &0.76 &\textcolor{Coral3}{0.68} &1\,018 &2\,591 &1\,449  &3\,293\\\hline
   \parbox[t]{1mm}{\multirow{2}{*}{\rotatebox[origin=c]{90}{\mixed}}} & \slm &0.83  &0.80 &0.84 &0.81 &1\,142 &2\,926 &1\,511 &3\,732\\
    & \orm &0.82 &0.80 &0.84 &0.83 &1\,159 &2\,944 &1\,555 &\textcolor{Coral3}{3\,746}\\
     \hline 
    \multicolumn{10}{|c|}{\claude}\\
    & & \multicolumn{4}{c?}{\textsc{Accuracy} $\alpha_{\text{MaxC}}$}   & \multicolumn{4}{c|}{\textsc{Total Tokens}} \\
  & & \multicolumn{2}{c|}{\textsc{\std}} & \multicolumn{2}{c?}{\textsc{CoT}}  & \multicolumn{2}{c|}{\textsc{\std}} & \multicolumn{2}{c|}{\textsc{CoT}} \\
  & \textbf{Modality}  & \textsc{Zero} &  \textsc{Few} & \textsc{Zero} &  \textsc{Few}  & \textsc{Zero} &  \textsc{Few} & \textsc{Zero} &  \textsc{Few} \\\hline
    & \txt &0.85 &0.77 &\textbf{0.99} &\textbf{0.99} &\textbf{310} &564 &1\,196  &1\,852  \\\hline
    \parbox[t]{1mm}{\multirow{2}{*}{\rotatebox[origin=c]{90}{\visual}}} &\slv &0.83 &0.88 &0.86 &0.86 &1\,571 &3\,579 &2\,355 &4\,589\\
    & \orv &0.76 &\textcolor{Coral3}{0.71} &0.73 &0.75 &1\,487 &3\,496 &2\,341 &4\,615\\\hline
   \parbox[t]{1mm}{\multirow{2}{*}{\rotatebox[origin=c]{90}{\mixed}}} & \slm &0.89  &0.84 &0.97 &\textbf{0.99} &1\,713 &3\,936 &2\,531 &\textcolor{Coral3}{5\,112}\\
    & \orm &0.80 &0.83 &0.97 &\textbf{0.99} &1\,631 &3\,854 &2\,436 &5\,015 \\\hline
    \end{tabular}}
    \vspace{1mm}
    \caption{Experiment 1: Performance on task \textsc{Max Clique}. \\Best (worst) values in \textbf{bold} (\textcolor{Coral3}{red}).}
    \label{tab:mc}
\end{table}

\renewcommand{\arraystretch}{1.8}
\newcolumntype{?}{!{\vrule width 1.5pt}}

\begin{table}[h]
\scriptsize
    \centering
    \resizebox{0.8\columnwidth}{!}{\begin{tabular}{|c c?c|c|c|c?c|c|c|c|}\hline
    \multicolumn{10}{|c|}{\gpt}\\
  & & \multicolumn{4}{c?}{\textsc{Accuracy} $\alpha_{\text{MinVC}}$}   & \multicolumn{4}{c|}{\textsc{Total Tokens}} \\
  & & \multicolumn{2}{c|}{\textsc{\std}} & \multicolumn{2}{c?}{\textsc{CoT}}  & \multicolumn{2}{c|}{\textsc{\std}} & \multicolumn{2}{c|}{\textsc{CoT}} \\
  & \textbf{Modality}  & \textsc{Zero} &  \textsc{Few} & \textsc{Zero} &  \textsc{Few}  & \textsc{Zero} &  \textsc{Few} & \textsc{Zero} &  \textsc{Few} \\\hline
    & \txt &0.74 &0.63 &0.77 &\textbf{0.78} &\textbf{353} &541 &1257  &1831  \\\hline
    \parbox[t]{1mm}{\multirow{2}{*}{\rotatebox[origin=c]{90}{\visual}}} &\slv &0.76 &0.77 &0.56 &0.62 &1\,017 &2\,580 &1\,415 &3\,374\\
    & \orv &0.63 &0.64 &\textcolor{Coral3}{0.53} &0.54 &1\,090 &2\,653 &1\,485  &3\,433\\\hline
   \parbox[t]{1mm}{\multirow{2}{*}{\rotatebox[origin=c]{90}{\mixed}}} & \slm &\textbf{0.78}  &0.70 &0.69 &0.66 &1\,204 &2\,931 &1\,605 &3\,855\\
    & \orm &0.70 &0.73 &0.73 &0.64 &1\,272 &2\,996 &1\,688 &\textcolor{Coral3}{3\,939}\\
    \hline
    \multicolumn{10}{|c|}{\claude}\\
    & & \multicolumn{4}{c?}{\textsc{Accuracy} $\alpha_{\text{MinVC}}$}   & \multicolumn{4}{c|}{\textsc{Total Tokens}} \\
  & & \multicolumn{2}{c|}{\textsc{\std}} & \multicolumn{2}{c?}{\textsc{CoT}}  & \multicolumn{2}{c|}{\textsc{\std}} & \multicolumn{2}{c|}{\textsc{CoT}} \\
  & \textbf{Modality}  & \textsc{Zero} &  \textsc{Few} & \textsc{Zero} &  \textsc{Few}  & \textsc{Zero} &  \textsc{Few} & \textsc{Zero} &  \textsc{Few} \\\hline
    & \txt &0.69 &\textcolor{Coral3}{0.60} &0.83 &0.82 &\textbf{353} &544 &1\,244  &1\,953  \\\hline
    \parbox[t]{1mm}{\multirow{2}{*}{\rotatebox[origin=c]{90}{\visual}}} &\slv &0.72 &0.72 &0.76 &0.64 &1\,560 &3\,409 &2\,608 &4\,608\\
    & \orv &0.70 &0.74 &0.76 &0.77 &1\,556 &3\,405 &2\,709 &4\,660\\\hline
   \parbox[t]{1mm}{\multirow{2}{*}{\rotatebox[origin=c]{90}{\mixed}}} & \slm &0.73  &0.66 &0.82 &0.81 &1\,752 &3\,766 &2\,847 &\textcolor{Coral3}{5\,069}\\
    & \orm &0.69 &0.67 &0.80 &0.81 &1\,745 &3\,763 &2\,691 &5\,024\\\hline
    \end{tabular}}
    \vspace{1mm}
    \caption{Experiment 1: Performance on task \textsc{Min Vertex Cover}. \\Best (worst) values in \textbf{bold} (\textcolor{Coral3}{red}).}
    \label{tab:mvc}
\end{table}

\begin{figure}
    \centering
    \includegraphics[width=0.6\linewidth]{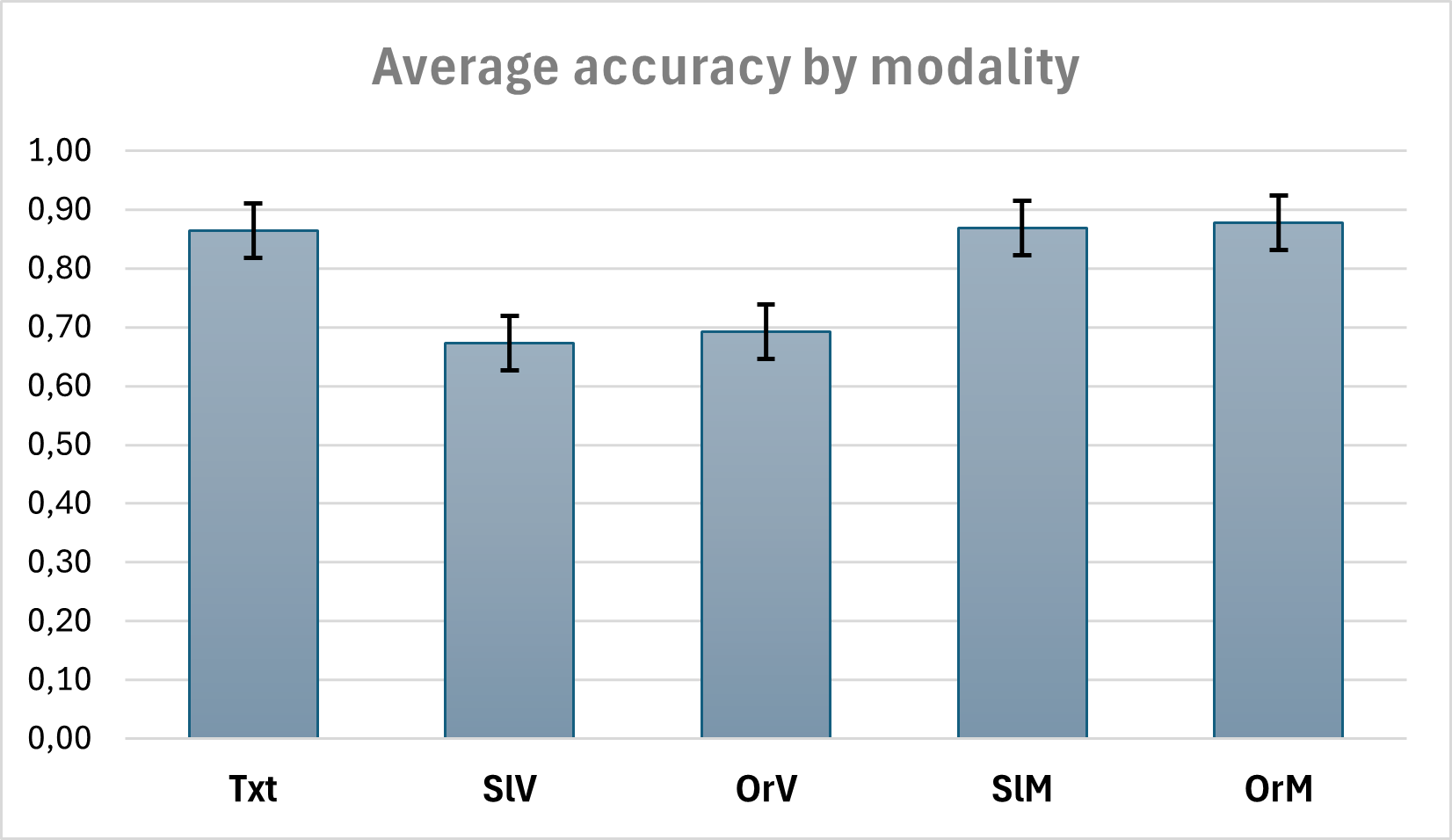}
    \caption{Experiment 1: Average accuracy by modality.}
    \label{fig:overall-acc-a}
\end{figure}

\begin{figure}
    \centering
    \includegraphics[width=0.6\linewidth]{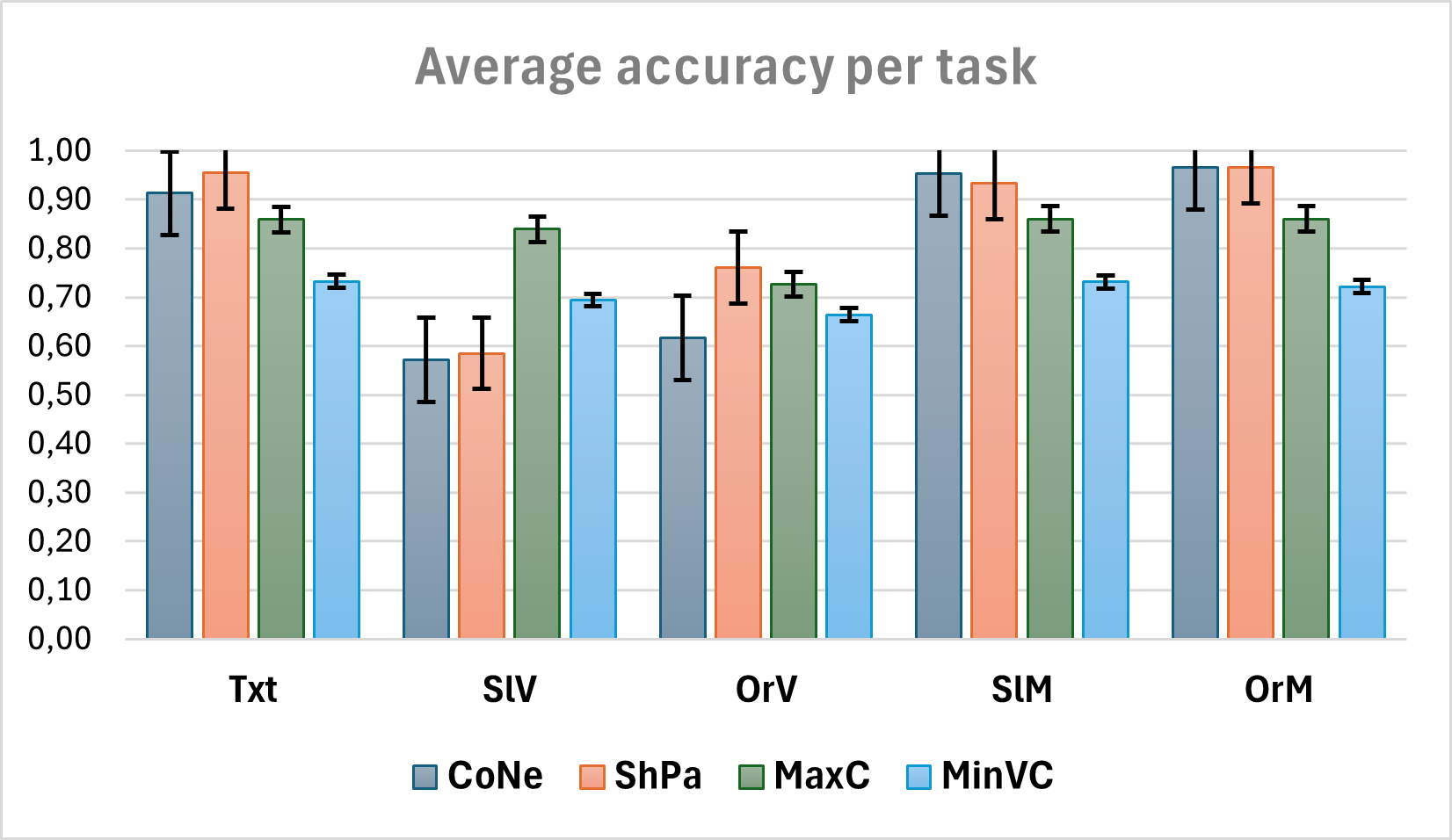}
    \caption{Experiment 1: Average accuracy by modality per task.}
    \label{fig:overall-acc-b}
\end{figure}

The results of \textbf{Experiment 1} are detailed in \Cref{tab:cn,tab:sp,tab:mc,tab:mvc}. The accuracy by modality averaged over all tasks and prompting techniques is shown in \Cref{fig:overall-acc-a}. It can be seen that, grouping the responses from \gpt and \claude together, \slm and \orm have an average accuracy of 0.87 and 0.88, respectively, slightly outperforming \txt (0.86), which in turn largely outperforms both \slv and \orv (0.67 and 0.69, respectively). Moreover, this trend is confirmed also when looking at each single LLM. Of particular interest with respect to \textbf{R1}, we observe that \orm (0.88) performs slightly better than \slm (0.87), and that, consistently, \orv (0.69) is better than \slv (0.67).  When analyzing the data separately per task (and over both LLMs), see \Cref{fig:overall-acc-b}, the above pattern is confirmed for task \textsc{CoNe}, and it is very prominent for task \textsc{ShPa}. In particular, \orm ($\alpha_{CoNe}=0.97$, $\alpha_{ShPa}=0.97$) performs better than \slm ($\alpha_{CoNe}=0.95$, $\alpha_{ShPa}=0.93$), while \orv ($\alpha_{CoNe}=0.62$, $\alpha_{ShPa}=0.76$) is better than \slv ($\alpha_{CoNe}=0.57$, $\alpha_{ShPa}=0.59$). On the other hand, the pattern is reversed for the more complex tasks \textsc{MaxC} and \textsc{MinVC}. Namely, \slm ($\alpha_{MaxC}=0.86$, $\alpha_{MinVC}=0.73$) performs equally or slightly better than \orm ($\alpha_{MaxC}=0.86$, $\alpha_{MinVC}=0.72$), while \slv ($\alpha_{MaxC}=0.84$, $\alpha_{MinVC}=0.69$) is better than \orv ($\alpha_{MaxC}=0.73$, $\alpha_{MinVC}=0.66$).

\begin{figure}
    \centering
    \includegraphics[width=0.6\linewidth]{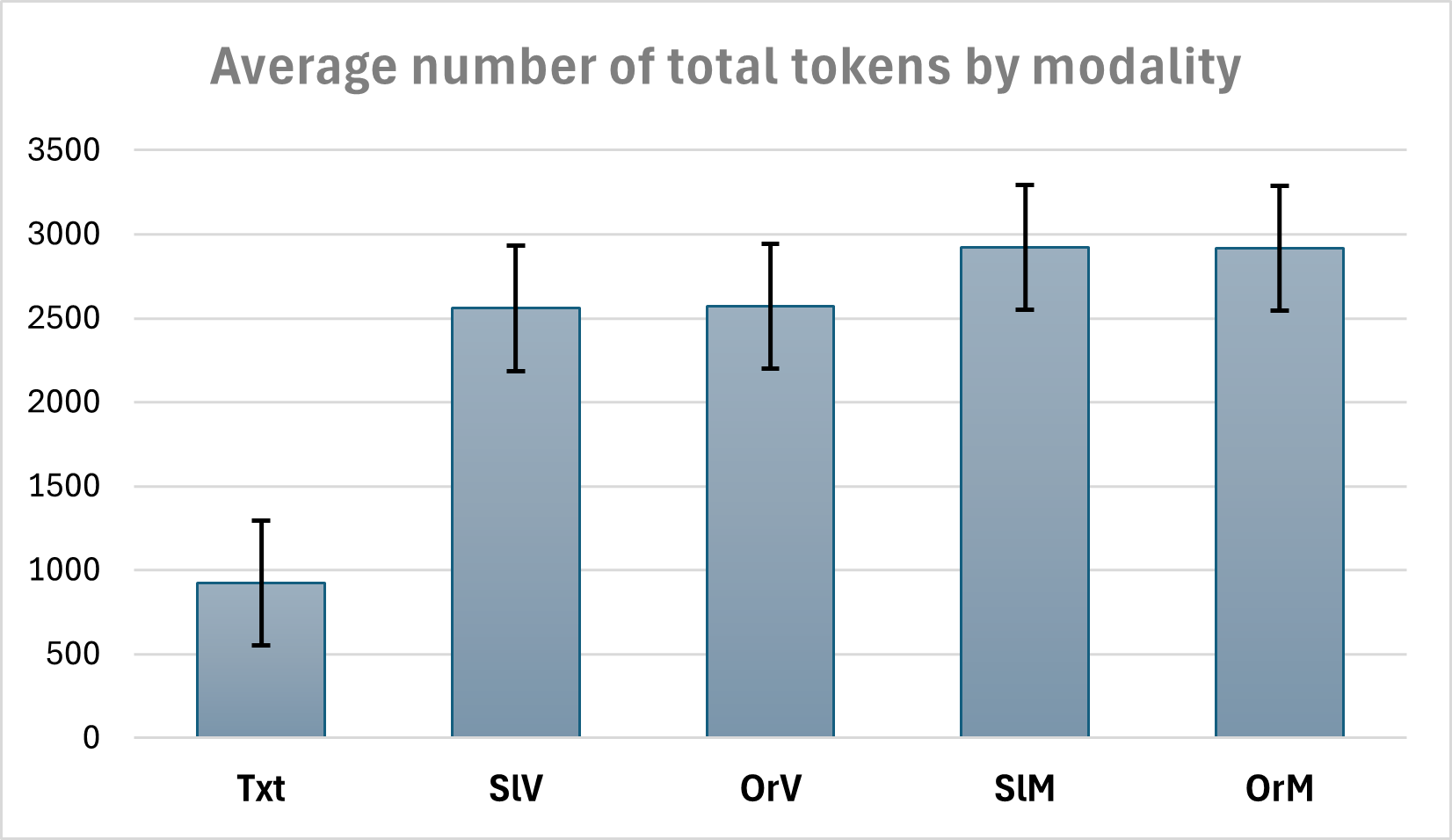}
    \caption{Experiment 1: Average number of total tokens by modality.}
    \label{fig:overall-lat-a}
\end{figure}

\begin{figure}
    \centering
    \includegraphics[width=0.6\linewidth]{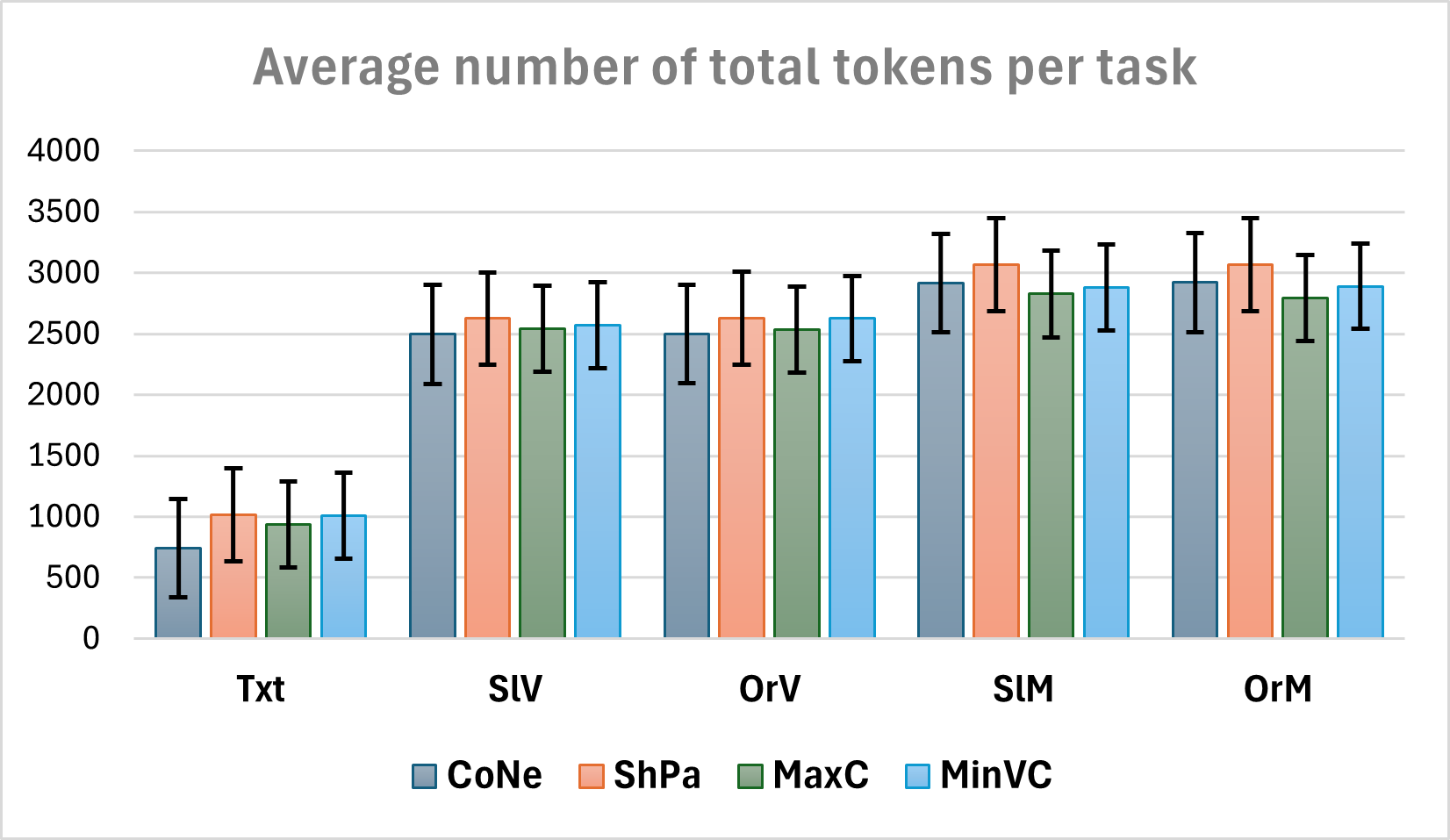}
    \caption{Experiment 1: Average number of total tokens by modality by task.}
    \label{fig:overall-lat-b}
\end{figure}

\medskip We conclude with a brief discussion about latency, measured in terms of total number of tokens. Average figures aggregated by modality are shown in \Cref{fig:overall-lat-a}. As expected, each \mixed modality requires a number of tokens that is about the number of tokens of \txt plus the number of tokens of the corresponding \vis modality. Also, it comes with no surprises that the two \vis modalities require about the same number of tokens. As we have used relatively small graphs, the \txt modality uses the least number of tokens; on the other hand, \vis modalities are expected to scale better for larger graphs (see also \cite{DBLP:conf/naacl/DasGSK24}). The trend does not change when analyzing the data separately per task, see \Cref{fig:overall-lat-b}, with more complex tasks requiring a slightly larger amount of tokens, mostly due to longer input prompts. 

\medskip
\noindent\textbf{Key finding for R1.} Our experiments reveal that the layout paradigm impacts the ability of the LLM to answer graph queries. In particular, orthogonal drawings appear superior on those tasks in which it is important to follow local connections or paths. This is probably due to the good readability of edges, which are drawn as orthogonal chains offering high vertex and crossing resolution. On the other hand, straight-line drawings have led to better results on more complex tasks involving the global structure of the graph. This can be justified by the fact that these drawings are produced by force-directed algorithms, which are good at highlighting symmetries and local structures (such as cliques).

\subsection{Experiment 2: Introducing a new prompting technique}

In the previous experiment, we noted that \txt performs better compared to \vis. This is not surprising due to the fact that inputs in the \txt modality contain the whole adjacency list of the graph. On the other hand, we have also seen that the \mix modality often leads to even better performance. Following this discussion, in this experiment we introduce and evaluate a new prompting technique, called \soall (\soal), whose goal is to drive the model through a reasoning strategy in which a preliminary extraction of the adjacency list of the graph from the image is performed, in order to enhance downstream tasks. The rationale is that this technique may trigger the LLM to produce a prompting that resembles the one used in the \mix modality. Obviously, the other side of the coin is that mistakes made by the LLM in extracting the adjacency list are likely to cause mistakes in the subsequent execution of the specific task.  Full prompts can be found in our public repository\footnote{To be provided after publication or under request.}.

\smallskip
\subsubsection{Experimental set-up}
For the sake of \textbf{Experiment 2}, we extend the set-up of \textbf{Experiment 1} by introducing the \soal prompting technique. The set of tasks and graph benchmarks is therefore the same, while the input modalities are restricted to \slv and \orv, since \soal applies only to images. 

\smallskip
\subsubsection{Results}

\begin{table*}[h]
\renewcommand{\arraystretch}{1.8}
\scriptsize
    \centering
    \begin{tabular}{|c c?c|c|c|c|c|c?c|c|c|c|c|c|}
    \hline 
    \multicolumn{14}{|c|}{\gpt}\\
  & & \multicolumn{6}{c?}{\textsc{Accuracy $\alpha_{\text{CoNe}}$}}   & \multicolumn{6}{c|}{\textsc{Total Tokens}} \\
  & & \multicolumn{2}{c|}{\textsc{Std}} & \multicolumn{2}{c|}{\textsc{SoAL}} & \multicolumn{2}{c?}{\textsc{CoT}}  & \multicolumn{2}{c|}{\textsc{Std}} & \multicolumn{2}{c|}{\textsc{SoAL}} & \multicolumn{2}{c|}{\textsc{CoT}} \\
  & \textbf{Modality}  & \textsc{Zero} &  \textsc{Few} & \textsc{Zero} &  \textsc{Few} & \textsc{Zero} &  \textsc{Few} &  \textsc{Zero} &  \textsc{Few}  & \textsc{Zero} &  \textsc{Few} & \textsc{Zero} &  \textsc{Few} \\\hline
    
    \parbox[t]{1mm}{\multirow{2}{*}{\rotatebox[origin=c]{90}{\visual}}} & \slv &0.60 &\textbf{0.63} & 0.59 & 0.59 &0.58 &0.55 &\textbf{969} &2\,557 & 1\,352 & \textcolor{Coral3}{3\,510} &1\,128 &2\,883\\
    & \orv &0.54 &0.60 & 0.57 & \textcolor{Coral3}{0.48} &0.56  &0.55 &1\,002 &2\,590 & 1\,340 & 3\,509 & 1\,162  & 2\,911\\\hline
   
\hline
\multicolumn{14}{|c|}{\claude}\\
  & & \multicolumn{6}{c?}{\textsc{Accuracy $\alpha_{\text{CoNe}}$}}   & \multicolumn{6}{c|}{\textsc{Total Tokens}} \\
  & & \multicolumn{2}{c|}{\textsc{Std}} & \multicolumn{2}{c|}{\textsc{SoAL}} & \multicolumn{2}{c?}{\textsc{CoT}}  & \multicolumn{2}{c|}{\textsc{Std}} & \multicolumn{2}{c|}{\textsc{SoAL}} & \multicolumn{2}{c|}{\textsc{CoT}} \\
  & \textbf{Modality}  & \textsc{Zero} &  \textsc{Few} &\textsc{Zero} &  \textsc{Few} &\textsc{Zero} &  \textsc{Few} & \textsc{Zero} &  \textsc{Few}  & \textsc{Zero} &  \textsc{Few} & \textsc{Zero} &  \textsc{Few} \\\hline
    \parbox[t]{1mm}{\multirow{2}{*}{\rotatebox[origin=c]{90}{\visual}}} &\slv &0.62 &\textcolor{Coral3}{0.35} & 0.63 & 0.66 &0.62 &0.62  &1\,540 &4\,298 & 1\,944 & \textcolor{Coral3}{5\,295} &1\,882 &4\,701\\
    & \orv &0.73 &\textcolor{Coral3}{0.35} & 0.73 & 0.77 &0.78 &\textbf{0.82}  &\textbf{1\,510} &4\,264 & 1\,912 & 5\,252 & 1\,852  & 4\,686\\\hline
    \end{tabular}
    \vspace{1mm}
    \caption{Experiment 2: Performance on task \textsc{Common Neighbor}. \\Best (worst) values in \textbf{bold} (\textcolor{Coral3}{red}).}
    \label{tab:cn-soal}
\end{table*}

\begin{table*}[h]
\renewcommand{\arraystretch}{1.8}
\scriptsize
    \centering
    \begin{tabular}{|c c?c|c|c|c|c|c?c|c|c|c|c|c|}
    \hline 
    \multicolumn{14}{|c|}{\gpt}\\
  & & \multicolumn{6}{c?}{\textsc{Accuracy $\alpha_{\text{ShPa}}$}}   & \multicolumn{6}{c|}{\textsc{Total Tokens}} \\
  & & \multicolumn{2}{c|}{\textsc{Std}} & \multicolumn{2}{c|}{\textsc{SoAL}} & \multicolumn{2}{c?}{\textsc{CoT}}  & \multicolumn{2}{c|}{\textsc{Std}} & \multicolumn{2}{c|}{\textsc{SoAL}} & \multicolumn{2}{c|}{\textsc{CoT}} \\
  & \textbf{Modality}  & \textsc{Zero} &  \textsc{Few} & \textsc{Zero} &  \textsc{Few} & \textsc{Zero} &  \textsc{Few} &  \textsc{Zero} &  \textsc{Few}  & \textsc{Zero} &  \textsc{Few} & \textsc{Zero} &  \textsc{Few} \\\hline
    
    \parbox[t]{1mm}{\multirow{2}{*}{\rotatebox[origin=c]{90}{\visual}}} & \slv & 0.49 &\textcolor{Coral3}{0.47} & 0.56 & 0.49 &0.55 &0.52 &\textbf{982} &2\,573 & 1\,445 & 3\,737 &1\,365 &3\,186\\
    & \orv &\textbf{0.71} &0.65 & 0.69 & 0.63 &0.69  &0.70 & 1\,020 & 2\,609 & 1\,542 & \textcolor{Coral3}{4\,199} & 1\,376  & 3\,217\\\hline
   
\hline
\multicolumn{14}{|c|}{\claude}\\
  & & \multicolumn{6}{c?}{\textsc{Accuracy $\alpha_{\text{ShPa}}$}}   & \multicolumn{6}{c|}{\textsc{Total Tokens}} \\
  & & \multicolumn{2}{c|}{\textsc{Std}} & \multicolumn{2}{c|}{\textsc{SoAL}} & \multicolumn{2}{c?}{\textsc{CoT}}  & \multicolumn{2}{c|}{\textsc{Std}} & \multicolumn{2}{c|}{\textsc{SoAL}} & \multicolumn{2}{c|}{\textsc{CoT}} \\
  & \textbf{Modality}  & \textsc{Zero} &  \textsc{Few} &\textsc{Zero} &  \textsc{Few} &\textsc{Zero} &  \textsc{Few} & \textsc{Zero} &  \textsc{Few}  & \textsc{Zero} &  \textsc{Few} & \textsc{Zero} &  \textsc{Few} \\\hline
    \parbox[t]{1mm}{\multirow{2}{*}{\rotatebox[origin=c]{90}{\visual}}} &\slv & 0.64 & 0.65 & 0.64 & \textcolor{Coral3}{0.63} &0.68 & 0.68  &1\,551 &4\,309 & 2\,116 & \textcolor{Coral3}{5\,592} & 2\,025 & 5\,014\\
    & \orv & 0.82 & 0.74 & 0.86 & 0.88 & 0.88 &\textbf{0.89}  & \textbf{1\,525} &4\,284 & 2\,091 & 5\,577 & 2\,004  & 4\,992\\\hline
    \end{tabular}
    \vspace{1mm}
    \caption{Experiment 2: Performance on task \textsc{Shortest Path}. \\Best (worst) values in \textbf{bold} (\textcolor{Coral3}{red}).}
    \label{tab:sp-soal}
\end{table*}

\begin{table*}[h]
\renewcommand{\arraystretch}{1.8}
\scriptsize
    \centering
    \begin{tabular}{|c c?c|c|c|c|c|c?c|c|c|c|c|c|}
    \hline 
    \multicolumn{14}{|c|}{\gpt}\\
  & & \multicolumn{6}{c?}{\textsc{Accuracy $\alpha_{\text{MaxC}}$}}   & \multicolumn{6}{c|}{\textsc{Total Tokens}} \\
  & & \multicolumn{2}{c|}{\textsc{Std}} & \multicolumn{2}{c|}{\textsc{SoAL}} & \multicolumn{2}{c?}{\textsc{CoT}}  & \multicolumn{2}{c|}{\textsc{Std}} & \multicolumn{2}{c|}{\textsc{SoAL}} & \multicolumn{2}{c|}{\textsc{CoT}} \\
  & \textbf{Modality}  & \textsc{Zero} &  \textsc{Few} & \textsc{Zero} &  \textsc{Few} & \textsc{Zero} &  \textsc{Few} &  \textsc{Zero} &  \textsc{Few}  & \textsc{Zero} &  \textsc{Few} & \textsc{Zero} &  \textsc{Few} \\\hline
    
    \parbox[t]{1mm}{\multirow{2}{*}{\rotatebox[origin=c]{90}{\visual}}} & \slv & 0.78 & 0.79 &  0.83 &  0.79 & 0.85 & \textbf{0.86} & \textbf{1\,002} &2\,574 & 1\,330 & 3\,336 & 1\,404 & 3\,255\\
    & \orv & 0.71 & 0.71 & \textcolor{Coral3}{0.65} & 0.69 & 0.76  & 0.68 & 1\,018 &2\,591 & 1\,337 & \textcolor{Coral3}{3\,367} & 1\,449  & 3\,293\\\hline
   
\hline
\multicolumn{14}{|c|}{\claude}\\
  & & \multicolumn{6}{c?}{\textsc{Accuracy $\alpha_{\text{MaxC}}$}}   & \multicolumn{6}{c|}{\textsc{Total Tokens}} \\
  & & \multicolumn{2}{c|}{\textsc{Std}} & \multicolumn{2}{c|}{\textsc{SoAL}} & \multicolumn{2}{c?}{\textsc{CoT}}  & \multicolumn{2}{c|}{\textsc{Std}} & \multicolumn{2}{c|}{\textsc{SoAL}} & \multicolumn{2}{c|}{\textsc{CoT}} \\
  & \textbf{Modality}  & \textsc{Zero} &  \textsc{Few} &\textsc{Zero} &  \textsc{Few} &\textsc{Zero} &  \textsc{Few} & \textsc{Zero} &  \textsc{Few}  & \textsc{Zero} &  \textsc{Few} & \textsc{Zero} &  \textsc{Few} \\\hline
    \parbox[t]{1mm}{\multirow{2}{*}{\rotatebox[origin=c]{90}{\visual}}} &\slv & 0.83 & \textbf{0.88} & 0.86 & 0.84 & 0.86 & 0.86  & 1\,571 & 3\,579 & 2\,155 & 4\,528 & 2\,355 & 4\,589\\
    & \orv & 0.76 & \textcolor{Coral3}{0.71} & 0.73 & 0.73 & 0.73 & 0.75  & \textbf{1\,487} & 3\,496 & 2\,066 & 4\,457  & 2\,341 & \textcolor{Coral3}{4\,615}\\\hline
    \end{tabular}
    \vspace{1mm}
    \caption{Experiment 2:  Performance on task \textsc{Max Clique}. \\Best (worst) values in \textbf{bold} (\textcolor{Coral3}{red}).}
    \label{tab:mc-soal}
\end{table*}

\begin{table*}[h]
\renewcommand{\arraystretch}{1.8}
\scriptsize
    \centering
   \begin{tabular}{|c c?c|c|c|c|c|c?c|c|c|c|c|c|}
    \hline 
    \multicolumn{14}{|c|}{\gpt}\\
  & & \multicolumn{6}{c?}{\textsc{Accuracy $\alpha_{\text{MinVC}}$}}   & \multicolumn{6}{c|}{\textsc{Total Tokens}} \\
  & & \multicolumn{2}{c|}{\textsc{Std}} & \multicolumn{2}{c|}{\textsc{SoAL}} & \multicolumn{2}{c?}{\textsc{CoT}}  & \multicolumn{2}{c|}{\textsc{Std}} & \multicolumn{2}{c|}{\textsc{SoAL}} & \multicolumn{2}{c|}{\textsc{CoT}} \\
  & \textbf{Modality}  & \textsc{Zero} &  \textsc{Few} & \textsc{Zero} &  \textsc{Few} & \textsc{Zero} &  \textsc{Few} &  \textsc{Zero} &  \textsc{Few}  & \textsc{Zero} &  \textsc{Few} & \textsc{Zero} &  \textsc{Few} \\\hline
    
    \parbox[t]{1mm}{\multirow{2}{*}{\rotatebox[origin=c]{90}{\visual}}} & \slv & 0.76 & \textbf{0.77} & 0.62 & 0.51 &0.56 &0.62 & \textbf{1\,017} & 2\,580 & 1\,487 & 3\,521 & 1\,415 & 3\,374\\
    & \orv & 0.63 &0.64 & \textcolor{Coral3}{0.49} & 0.53 &0.53  &0.54 & 1\,090 & 2\,653 & 1\,564 & \textcolor{Coral3}{3\,600} & 1\,485  & 3\,433\\\hline
   
\hline
\multicolumn{14}{|c|}{\claude}\\
  & & \multicolumn{6}{c?}{\textsc{Accuracy $\alpha_{\text{MinVC}}$}}   & \multicolumn{6}{c|}{\textsc{Total Tokens}} \\
  & & \multicolumn{2}{c|}{\textsc{Std}} & \multicolumn{2}{c|}{\textsc{SoAL}} & \multicolumn{2}{c?}{\textsc{CoT}}  & \multicolumn{2}{c|}{\textsc{Std}} & \multicolumn{2}{c|}{\textsc{SoAL}} & \multicolumn{2}{c|}{\textsc{CoT}} \\
  & \textbf{Modality}  & \textsc{Zero} &  \textsc{Few} &\textsc{Zero} &  \textsc{Few} &\textsc{Zero} &  \textsc{Few} & \textsc{Zero} &  \textsc{Few}  & \textsc{Zero} &  \textsc{Few} & \textsc{Zero} &  \textsc{Few} \\\hline
    \parbox[t]{1mm}{\multirow{2}{*}{\rotatebox[origin=c]{90}{\visual}}} &\slv & 0.72 & 0.72 & 0.67 & \textcolor{Coral3}{0.61} & 0.76 & 0.64 & 1\,560 & 3\,409 & 2\,214 & 4\,570 & 2\,608 & 4\,608\\
    & \orv &0.70 & 0.74 & 0.65 & 0.69 & 0.76 & \textbf{0.77} &\textbf{1\,556} & 3\,405 & 2\,171 & 4\,553 & 2\,709  & \textcolor{Coral3}{4\,660}\\\hline
    \end{tabular}
    \vspace{1mm}
    \caption{Experiment 2:  Performance on task \textsc{Min Vertex Cover}. \\Best (worst) values in \textbf{bold} (\textcolor{Coral3}{red}).}
    \label{tab:mvc-soal}
\end{table*}

\begin{figure}
    \centering
    \includegraphics[width=0.6\linewidth]{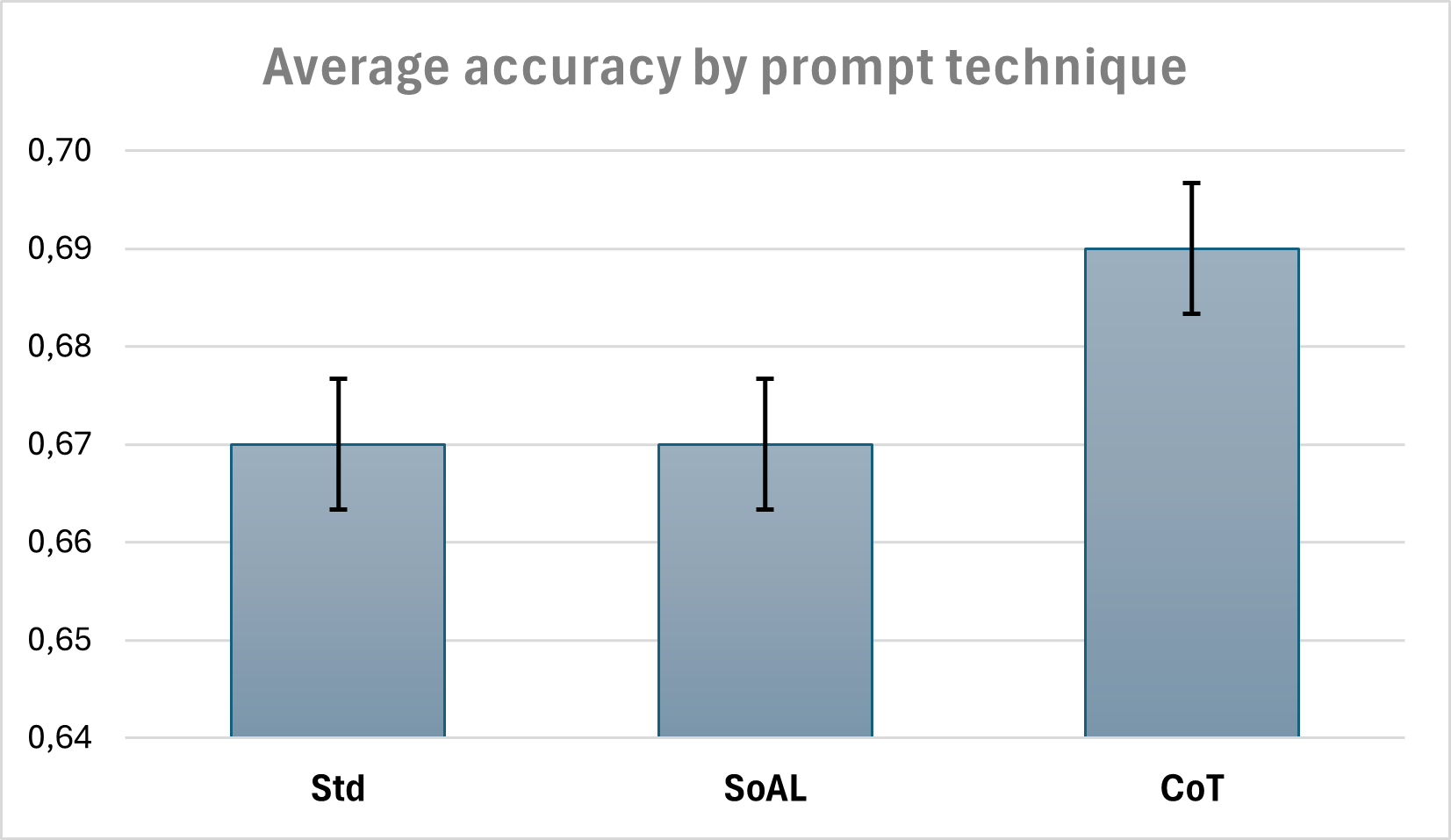}
    \caption{Experiment 2: Average accuracy by prompting technique.}
    \label{fig:overall-soal-acc-a}
\end{figure}

\begin{figure}
    \centering
    \includegraphics[width=0.6\linewidth]{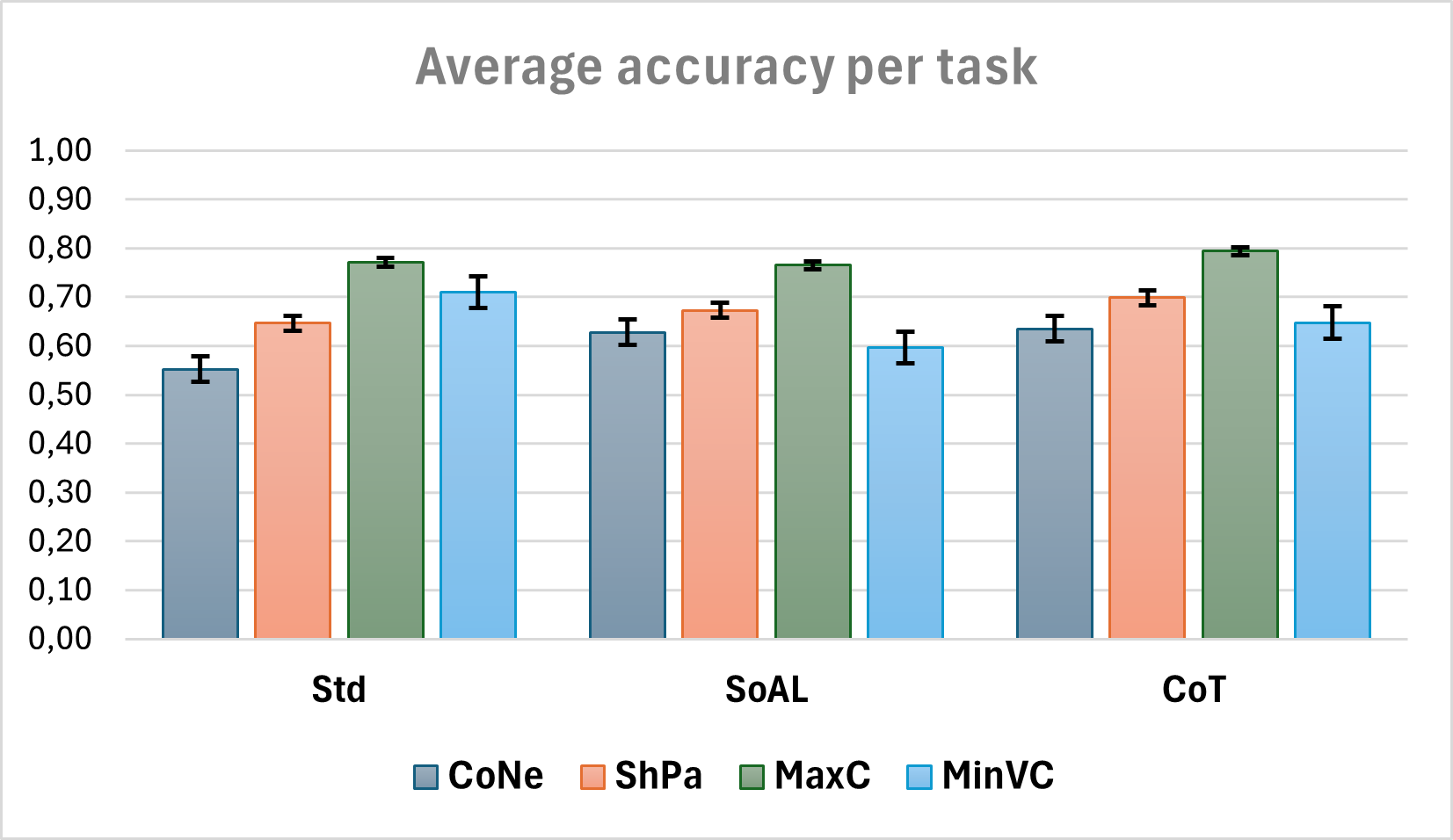}
    \caption{Experiment 2: Average accuracy by prompting technique per task.}
    \label{fig:overall-soal-acc-b}
\end{figure}

The results of \textbf{Experiment 2} are detailed in \Cref{tab:cn-soal,tab:sp-soal,tab:mc-soal,tab:mvc-soal}. The accuracy by prompting technique averaged over all tasks and input modalities is shown in \Cref{fig:overall-soal-acc-a}. It can be seen that, grouping the responses from \gpt and \claude together, \std and \soal lead to the same performance (0.67) while \cots performs slightly better (0.69). However, this trend is not consistent between the two LLMs; \gpt reveals overall better performance with \std (0.66), follow by \cots (0.63) and \soal (0.61), while \claude follows an opposite patterns, as it works better with \cots (0.76), followed by \soal (0.72) and \std (0.69).  When analyzing the data separately per task (and over both LLMs), see \Cref{fig:overall-soal-acc-b}, we have that for task \textsc{CoNe}, \cots ($\alpha_{CoNe}=0.64$) shows slightly better performance than \soal ($\alpha_{CoNe}=0.63$), which in turn behaves much better than \std  ($\alpha_{CoNe}=0.55$). The same pattern is confirmed  for task \textsc{ShPa}, with \cots ($\alpha_{ShPa}=0.70$) better than \soal ($\alpha_{ShPa}=0.67$), which in turn is better than \std ($\alpha_{ShPa}=0.65$). For task \textsc{MaxC}, \std and \soal show the same overall performance ($\alpha_{MaxC}=0.77$), while \cots works slightly better ($\alpha_{MaxC}=0.79$). On the opposite, for task \textsc{MinVC} we see that \std ($\alpha_{MinVC}=0.71$) works better than \cots ($\alpha_{MinVC}=0.65$), which is better than \soal ($\alpha_{MinVC}=0.60$).

\begin{figure}
    \centering
    \includegraphics[width=0.6\linewidth]{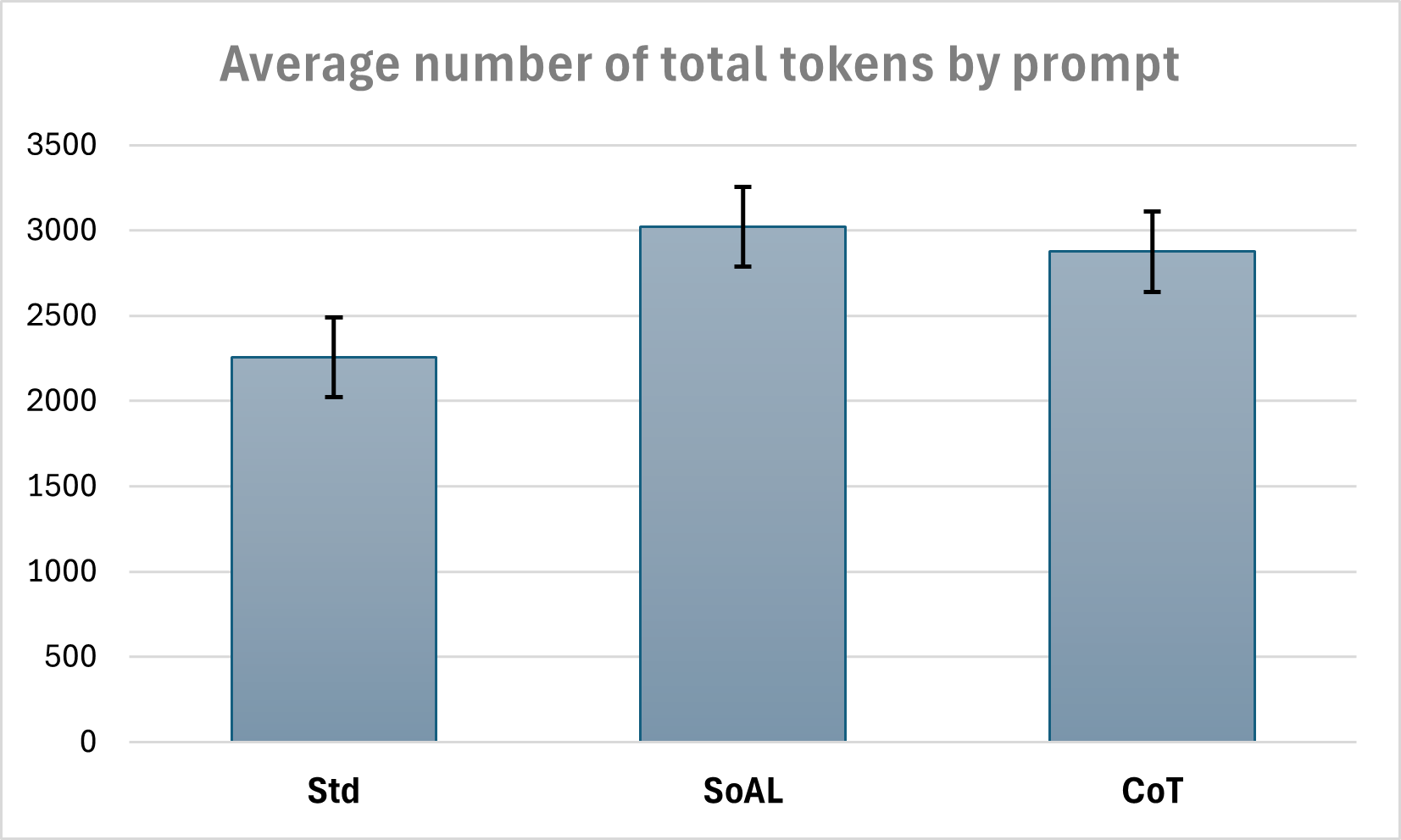}
    \caption{Experiment 2: Average number of total tokens by prompting technique.}
    \label{fig:overall-prompt-lat-a}
\end{figure}

\begin{figure}
    \centering
    \includegraphics[width=0.6\linewidth]{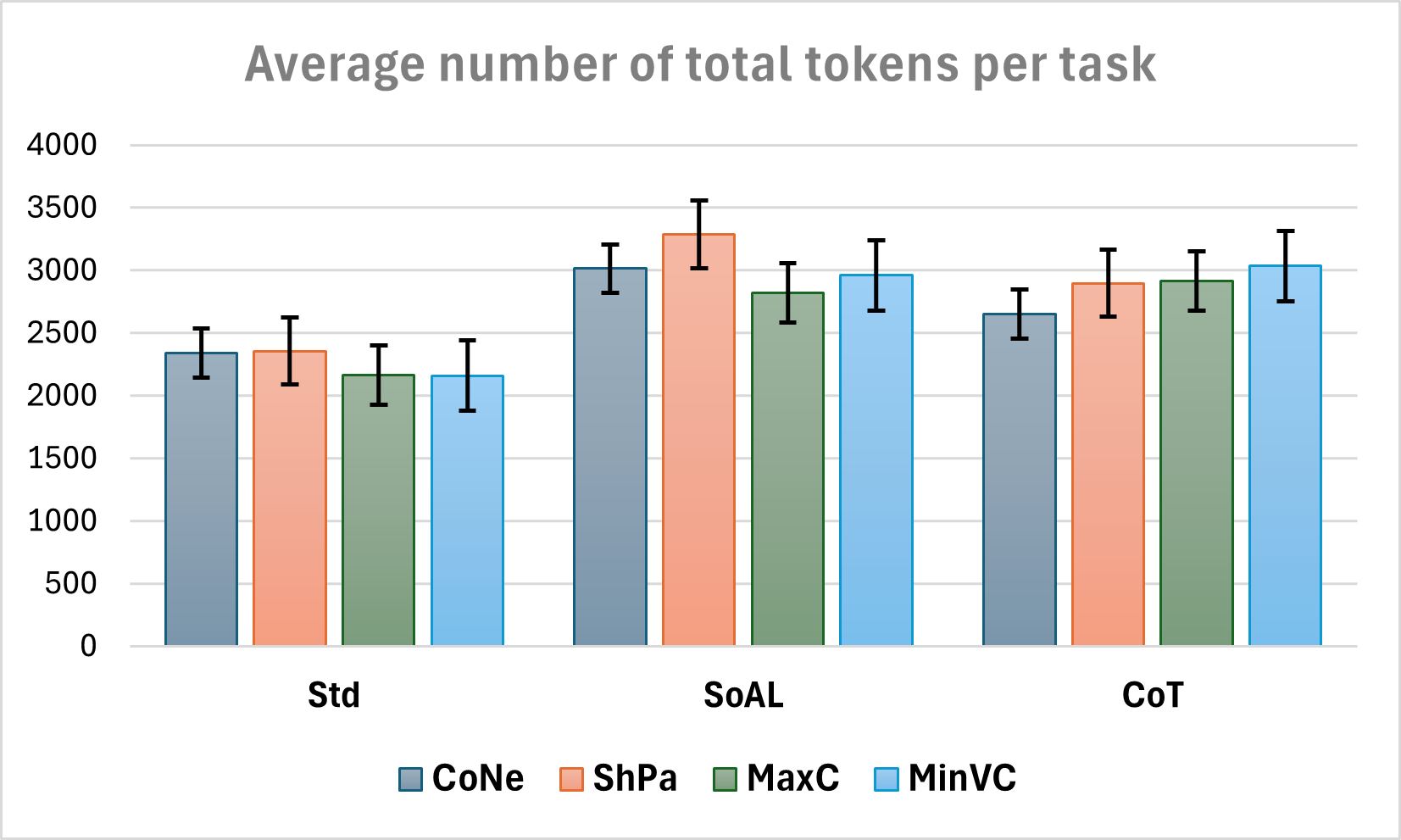}
    \caption{Experiment 2: Average number of total tokens by prompting technique per task.}
    \label{fig:overall-prompt-lat-b}
\end{figure}

\medskip We again conclude with a brief discussion about the total number of tokens. Average figures aggregated by prompting technique are shown in \Cref{fig:overall-prompt-lat-a}. We observe that \std requires the least number of tokens, while \cots and \soal require more tokens, namely +28\% and +33\%, respectively. On the other hand, \soal requires only +5\% additional tokens compared to \cots, even though it implies having the full adjacency list in output. Even more, when analyzing the data separately per task, see \Cref{fig:overall-prompt-lat-b}, we have that \std is still the more efficient technique, whereas \soal slightly overcomes \cots on \textsc{MaxC} and \textsc{MinVC}. By inspecting the outputs, we have seen that indeed complex tasks cause very long chains of thoughts.

\medskip
\noindent\textbf{Key finding for R2.} Our experiments do not reveal an overall better prompting technique. The newly introduced \soal technique is promising when used with \claude, especially for complex tasks for which the number of tokens is comparable with that of \cots. More in general, based on our results, experimenting different prompting techniques is advisable.

\subsection{Experiment 3: Evaluating the impact of improved quality metrics}

\subsubsection{Experimental set-up} For this experiment, the idea is to have a larger benchmark of graphs (described below), on which we first compute straight-line drawings (\slv), and then we manually improve the quality of such drawings obtaining a new modality (\sliv). We considered a single task, \textsc{ShPa}, since it represents a task balancing local and global exploration and hence requiring a good mix of local and global readability. Below are the novel elements of this experiment.

\begin{itemize}
   \item \textsc{Graph Benchmark 4} (\textsc{Bench-4}): $28$ graphs, with number of vertices between $7$ and $50$, with different topologies from small planar graphs to more complex graphs with dense communities. 
   \item \slivisual (\sliv): Straight-line drawings obtained by manually adjusting those obtained with a force-directed algorithm (\slv). The manual optimization is based on human experience and well-accepted metrics such as symmetry and number of edge crossings (see, e.g.,~\cite{DBLP:journals/jgaa/PurchaseAC02,DBLP:journals/jea/PurchaseCJ97}).
   \item \slimixed (\slim): It combines \txt with \sliv.
\end{itemize}

\subsubsection{Results}

\newcolumntype{?}{!{\vrule width 1.5pt}}
\begin{table}[h]
\renewcommand{\arraystretch}{1.8}
\scriptsize
    \centering
    \resizebox{0.8\columnwidth}{!}{\begin{tabular}{|c c?c|c|c|c?c|c|c|c|}
    \hline 
    \multicolumn{10}{|c|}{\gpt}\\
  & & \multicolumn{4}{c?}{\textsc{Accuracy}  $\alpha_{\text{ShPa}}$}   & \multicolumn{4}{c|}{\textsc{Total Tokens}} \\
  & & \multicolumn{2}{c|}{\textsc{Std}} & \multicolumn{2}{c?}{\textsc{CoT}}  & \multicolumn{2}{c|}{\textsc{Std}} & \multicolumn{2}{c|}{\textsc{CoT}} \\
  & \textbf{Modality}  & \textsc{Zero} &  \textsc{Few} & \textsc{Zero} &  \textsc{Few}  & \textsc{Zero} &  \textsc{Few} & \textsc{Zero} &  \textsc{Few} \\\hline
    \parbox[t]{1mm}{\multirow{2}{*}{\rotatebox[origin=c]{90}{\visual}}} &\slv &0.45 &0.41 &0.53 &0.54 &983 &2574 &1422 &3208\\
    & \sliv &0.49 &0.54 &0.54 &0.59 &983 &2574 &1392  &3214\\\hline
   \parbox[t]{1mm}{\multirow{2}{*}{\rotatebox[origin=c]{90}{\mixed}}} & \slm &0.84  &0.82 &0.95 &0.90 &1254 &3248 &1669 &3938\\
    & \slim &0.82 &0.83 &0.94 &0.91 &1254 &3248 &1653 &4011\\
    \hline 
    \multicolumn{10}{|c|}{\claude}\\
     & & \multicolumn{4}{c?}{\textsc{Accuracy} $\alpha_{\text{ShPa}}$}   & \multicolumn{4}{c|}{\textsc{Total Tokens}} \\
  & & \multicolumn{2}{c|}{\textsc{Std}} & \multicolumn{2}{c?}{\textsc{CoT}}  & \multicolumn{2}{c|}{\textsc{Std}} & \multicolumn{2}{c|}{\textsc{CoT}} \\
  & \textbf{Modality}  & \textsc{Zero} &  \textsc{Few} & \textsc{Zero} &  \textsc{Few}  & \textsc{Zero} &  \textsc{Few} & \textsc{Zero} &  \textsc{Few} \\\hline
    \parbox[t]{1mm}{\multirow{2}{*}{\rotatebox[origin=c]{90}{\visual}}} &\slv &0.56 &0.51 &0.55 &0.60 &1544 &4302 &2113 &5086\\
    & \sliv &0.66 &0.55 &0.64 &0.61 &1545 &4303 &2148  &5057\\\hline
   \parbox[t]{1mm}{\multirow{2}{*}{\rotatebox[origin=c]{90}{\mixed}}} & \slm &0.87  &0.51 &0.96 &0.78 &1763 &4924 &2468 &5822\\
    & \slim &0.84 &0.61 &0.96 &0.87 &1726 &4885 &2430 &5836 \\\hline
    \end{tabular}}
    \vspace{1mm}
    \caption{Experiment 3: Performance on task \textsc{Shortest Path}. \\Best (worst) values in \textbf{bold} (\textcolor{Coral3}{red}).}
    \label{tab:sp-improved}
\end{table}

\begin{figure}
    \centering
    \includegraphics[width=0.6\linewidth]{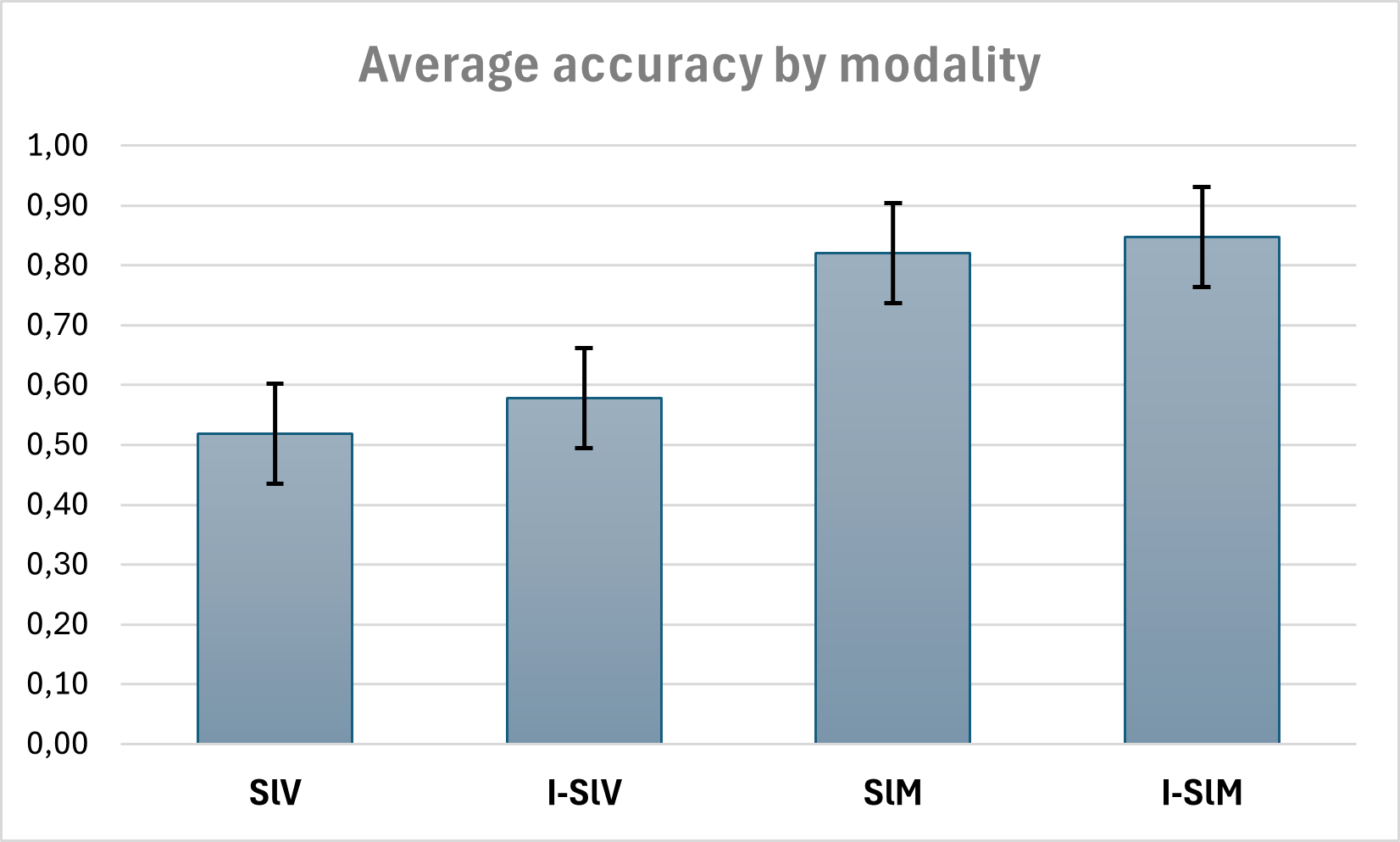}
    \caption{Experiment 3: Average accuracy by modality.}
    \label{fig:overall-imp-acc-a}
\end{figure}

The results of \textbf{Experiment 3} are detailed in \Cref{tab:sp-improved}. The overall accuracy by modality  is shown in \Cref{fig:overall-imp-acc-a}. Notably, the modality \sliv with improved drawings ($\alpha_{\textsc{ShPa}}=0.58$) performs better than \slv ($\alpha_{\textsc{ShPa}}=0.52$), and, similarly, \slim  ($\alpha_{\textsc{ShPa}}=0.85$) performs better than \slm ($\alpha_{\textsc{ShPa}}=0.82$). This trend is confirmed when analyzing \gpt and \claude separately, thus improved drawings appear superior irrespectively of the LLM.

\begin{figure}
    \centering
    \includegraphics[width=0.6\linewidth]{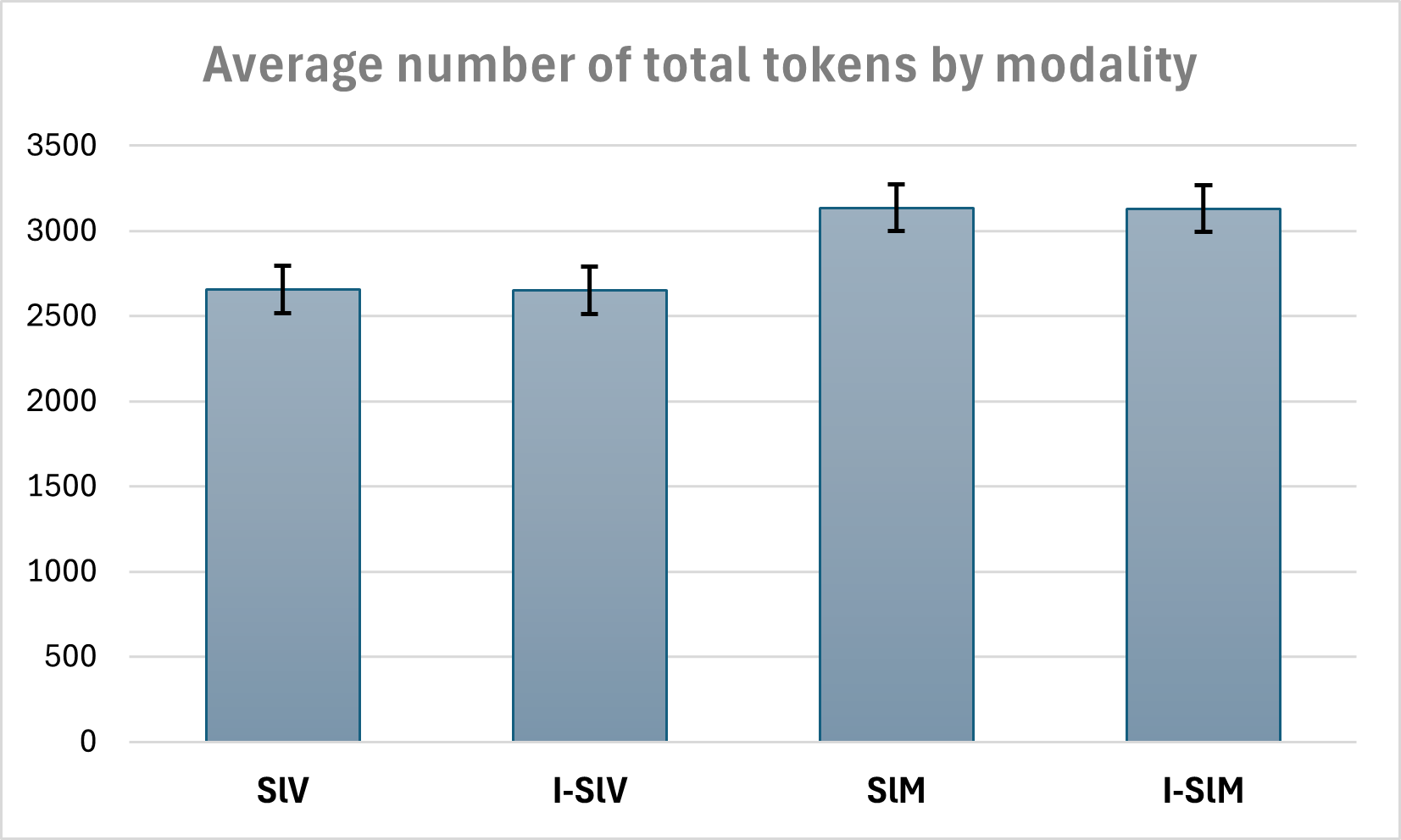}
    \caption{Experiment 3: Average number of total tokens by modality.}
    \label{fig:overall-imp-lat-a}
\end{figure}

In terms of latency, as shown in \Cref{fig:overall-imp-lat-a}, \sliv and \slv require about the same number of tokens, and the same holds for \slim and \slm. This is expected since small modifications to the image should not affect the number of tokens.

\medskip
\noindent\textbf{Key finding for R3.} Our experiments support the fact that improving the readability of a graph drawing based on human readability metrics increases the LLM's ability to solve tasks on the input graph. 

\section{Conclusions}\label{se:conclusions}

Our experiments shed light on the impact of graph layout paradigms, prompting techniques, and readability metrics on the ability of a LLM to solve graph-related tasks. We provided experimental evidence that carefully choosing the right layout paradigm and prompting technique based on the task to be executed, as well as optimizing the readability of the graph layout based on human feedback can significantly boost the LLM's performance. Our findings pave the way for new research that can leverage the adoption of AI assistants on a diverse range of tasks that exploit graphs as a data model. 

\subsection{Summary of key findings}

Three key findings, related to the three research questions proposed in \Cref{se:introduction}, are summarized below.

\begin{itemize}
    \item Orthogonal drawings, possibly thanks to their high angular resolution, lead to better performance compared to straight-line drawings on those tasks in which it is important to follow local connections or paths. On the other hand, straight-line drawings better unfold the graph structure and have led to better results on complex tasks requiring a more global understanding of the graph. 

    \item Since different LLMs may exhibit different behaviors when queried with the same prompting technique, it is advisable to experiment with different prompting techniques. On the other hand, our experiments confirm that \cotl is an effective strategy to obtain more accurate outputs (at the expenses of an increased latency and cost). In addition, our newly introduced \soall technique is promising when used with \claude, especially for complex tasks for which the number of tokens is comparable to that of \cotl. 

    \item Finally, our experiments support the idea that  quality metrics that typically impact on human readability of graph drawings  also impact on machine readability. Thus, applying a further step of manual optimization on a graph drawing may lead to better results in terms of accuracy.
\end{itemize}

\subsection{Limitations and future research directions}
In order to carry out our extensive experimental analysis we made $8,320$ calls to the APIs of OpenAI and Anthropic. However, there are still important limits that should be considered when generalizing our results beyond the experimental set-up. We conclude by summarizing such limits and by proposing future research directions that stem from our research.

\begin{itemize}

    \item \textbf{Large Language Models}.
    \begin{itemize}
        \item We tested two LLMs, which are state-of-the-art models at the time of writing. While other models may exhibit different behaviors, the overall consistency of the two considered models indicates a good robustness of our findings. Clearly, future models are likely to lead to better performance, and in particular the gap between \txt and \visual may be reduced.
        \item Our analysis focuses on a black-box approach that relies solely on foundational models. Designing integrated frameworks such as~\cite{DBLP:conf/nips/WeiFJZZWK024} is also a prominent option, which, however, may not be ideal for general-purpose AI agents.
    \end{itemize}
    \item \textbf{Graph benchmarks and tasks}.
    \begin{itemize}
        \item We chose our graphs by controlling their size and their structural properties (based on the specific task). This ensures that our benchmarks contain both simpler and harder instances with different scales. We believe that the performance of the visual modalities on local tasks stay stable on larger graphs, as the model still needs to analyze small patches of the overall image. On the other hand, the performance on global tasks is likely to degrade on larger and more complex graphs. Indeed, we expect that complex graphs lead to drawings with cluttered areas in which it is difficult to trace connections. 
        \item We did not consider directed graphs, for which specific graphic features indicating edge directions may impact the ability of the LLM to solve the given task. Experiments in this direction would be very interesting.
         \item We only considered structural tasks. Other tasks, requiring for instance node classification or link prediction are definitely worthy of attention.
    \end{itemize}
    
    \item \textbf{Readability metrics, layout paradigms, and graphical features}.
    \begin{itemize}
        \item We compared two popular drawing paradigms. Based on our experience, straight-line and orthogonal drawings cover a large fraction of potential use cases. It would be of interest to experiment other paradigms, such as polyline drawings or bundled drawings (see, e.g.,~\cite{DBLP:journals/tvcg/WallingerAANP22}).
    
        \item We disregarded other graphical features, such as, for instance, colors and shape. It would be of great interest to design experiments aimed at identifying the best graphical features, possibly in combination with the layout paradigm.
        
        \item In \textbf{Experiment 3}, we improved the drawings based on our own experience and standard readability metrics. Our results indicate that this is a promising direction. A more systematic study of what readability metrics have a greater impact on the LLM's abilities would provide additional insights. In particular, can LLMs be used as a reliable proxy for human-based experiments on the readability of graph drawings?
    \end{itemize}

\end{itemize}

\bibliographystyle{abbrv}
\bibliography{bibliography}

\end{document}